\documentclass[a4paper,fleqn]{cas-dc}

\usepackage{adjustbox}
\usepackage{algorithmic}
\usepackage{amsmath,amsfonts, amssymb}
\usepackage[authoryear]{natbib}
\usepackage{array}
\usepackage{balance}
\usepackage{booktabs}
\usepackage[british]{babel}
\usepackage[caption=false,font=normalsize,labelfont=sf,textfont=sf]{subfig}
\usepackage{colortbl}
\usepackage{commath}
\usepackage{gensymb}
\usepackage{graphicx}
\usepackage{hyperref}
\usepackage{lineno}
\usepackage{makecell}
\usepackage{mathtools}
\usepackage{multicol}
\usepackage{multirow}
\usepackage{placeins}
\usepackage{rotating}
\usepackage[shortlabels]{enumitem}
\usepackage{siunitx}
\usepackage{tabularx}
\usepackage{textcomp}
\usepackage[flushleft]{threeparttable}
\usepackage{stfloats}
\usepackage{url}
\usepackage{verbatim}
\usepackage{xcolor}

\setcellgapes{4pt}          
\newcommand\numberthis{\addtocounter{equation}{1}\tag{\theequation}}

\DeclarePairedDelimiter\floor{\lfloor}{\rfloor}

\def\tsc#1{\csdef{#1}{\textsc{\lowercase{#1}}\xspace}}
\tsc{WGM}
\tsc{QE}
\tsc{EP}
\tsc{PMS}
\tsc{BEC}
\tsc{DE}

\begin{document}
\let\WriteBookmarks\relax
\def\floatpagepagefraction{1}
\def\textpagefraction{.001}

\shorttitle{Classification of grapevine varieties using UAV hyperspectral imaging}    

\shortauthors{A. López, C. J. Ogayar, F. R. Feito, J. Sousa}  

\title [mode = title]{Classification of grapevine varieties using UAV hyperspectral imaging}  

%

\affiliation[1]{organization={Department of Computer Science},
            addressline={Campus Las Lagunillas s/n}, 
            city={Jaén},
            postcode={23071}, 
            country={Spain}}
            
\affiliation[2]{organization={Department of Software Engineering},
            city={Granada},
            postcode={18071}, 
            country={Spain}}
            
\affiliation[3]{organization={Centre for Robotics in Industry and Intelligent Systems (CRIIS), INESC Technology and Science},
            city={Porto},
            postcode={4200-465}, 
            country={Portugal}}
            
\affiliation[4]{organization={University of Trás-os-Montes e Alto Douro},
            city={Vila Real},
            postcode={5000-801}, 
            country={Portugal}}

\author[1]{Alfonso López}[orcid=0000-0003-1423-9496]
\cormark[0]
\ead{allopezr@ujaen.es}

\author[1]{Carlos J. Ogayar}[orcid=0000-0003-0958-990X]
\ead{cogayar@ujaen.es}
\author[1]{Francisco R. Feito}[orcid=0000-0001-8230-6529]
\ead{ffeito@ujaen.es}
\author[3,4]{Joaquim J. Sousa}[orcid=0000-0003-4533-930X]
\ead{jjsousa@utad.pt}

\begin{abstract}
The classification of different grapevine varieties is a relevant phenotyping task in Precision Viticulture since it enables estimating the growth of vineyard rows dedicated to different varieties, among other applications concerning the wine industry. This task can be performed with destructive methods that require time-consuming tasks, including data collection and analysis in the laboratory. However, Unmanned Aerial Vehicles (UAV) provide a more efficient and less prohibitive approach to collecting hyperspectral data, despite acquiring noisier data. Therefore, the first task is the processing of these data to correct and downsample large amounts of data. In addition, the hyperspectral signatures of grape varieties are very similar. In this work, a Convolutional Neural Network (CNN) is proposed for classifying seventeen varieties of red and white grape variants. Rather than classifying single samples, these are processed together with their neighbourhood. Hence, the extraction of spatial and spectral features is addressed with 1) a spatial attention layer and 2) Inception blocks. The pipeline goes from processing to dataset elaboration, finishing with the training phase. The fitted model is evaluated in terms of response time, accuracy and data separability, and compared with other state-of-the-art CNNs for classifying hyperspectral data. Our network was proven to be much more lightweight with a reduced number of input bands, a lower number of trainable weights and therefore, reduced training time. Despite this, the evaluated metrics showed much better results for our network ($\sim$99\% overall accuracy), in comparison with previous works barely achieving 81\% OA.

The source code is available at \verb|https://github.com/AlfonsoLRz/VineyardUAVClassification|.
\end{abstract}




\maketitle

\section{Introduction}
\label{sec:introduction}

Understanding vegetation development is a crucial aspect of crop management that impacts the effectiveness and productivity of agricultural efforts. Precision Agriculture (PA) involves observing agricultural variables
that affect crop production, which enables accounting for spatial and temporal variations, resulting in enhanced crop performance, reduced costs, and improved sustainability. Additionally, it provides a forecasting tool to accurately supply crop needs, such as water and nutrients. If applied to vines, this concept is known as Precision Viticulture (PV), with a wide variety of applications, ranging from the detection of biomass \citep{di_gennaro_evaluation_2020}, water content \citep{santesteban_high-resolution_2017, gutierrez_assessing_2021} and other compounds \citep{peng_prediction_2022}, to vigour estimation \citep{bramley_12_2010, campos_development_2019}, detection of plant diseases, pest surveillance \citep{mendes_vineinspector_2022}, analysis of grape maturity \citep{soubry_monitoring_2017} or yield estimation \citep{hassanzadeh_broadacre_2021}. They are either focused on grape clusters, leaves, stems or vineyard support \citep{singh_bibliometric_2022}. 

To achieve this goal, data is gathered through remote and proximal sensors for subsequent analysis. Remote Sensing (RS) data primarily includes aerial images captured by sensors attached to three main platforms: satellites, manned aircraft, and unmanned aerial vehicles (UAVs). Among these, satellite and UAV platforms are more extensively utilized, with UAVs gaining popularity due to their ability to improve both spatial and temporal resolution. There are several limitations associated with satellite platforms, such as high costs, low spatial resolution, and extended periods between revisits. Landsat 8, for example, captures panchromatic, multispectral, and thermal bands with spatial resolutions of 15 \si{\meter}, 30 \si{\meter} and 100 \si{\meter} \citep{ammoniaci_state_2021}, respectively, whereas its revisiting time is 8 days. In contrast, UAVs offer lower acquisition costs, the ability to integrate multiple sensors, and improved spatial resolution based on flight altitude. Consequently, these advantages make UAVs a suitable choice for contemporary PV.

The characterization of vineyard plots using UAVs is particularly challenging due to their variability, regarding the tree structure, inter-row spacing and surrounding elements (bare soil, shadowed areas, grassing, etc.). Therefore, high-detailed images are relevant for discriminating vegetation, soil and weeds, which have been previously shown to affect grape estimations \citep{ammoniaci_state_2021, sassu_advances_2021}. Consequently, a significant effort of previous studies is oriented toward the segmentation of the canopy \citep{padua_vineyard_2022}. In this regard, UAVs help to support make-decision systems by gathering precise information that enables the estimation of biophysical and performance-related features \citep{bramley_12_2010}.

This study examines hyperspectral data for classifying a significant number of grapevine varieties. First, data are explored to shed some light on the initial clustering of hyperspectral signatures of these varieties, and then, the feature reduction problem is studied over them. Once the shortcomings of working over these data are presented, a Deep Learning training pipeline is presented for providing a phenotyping tool applied to vineyards. The study's findings aim to demonstrate the classification capabilities of UAV hyperspectral data. In comparison with previous work, the proposed network is proven to perform well over UAV and satellite-based imagery.

\section{Related work}
\label{relatedwork}

The purpose of this section is to offer an overview of the research on hyperspectral data, encompassing both
conventional and novel approaches. Given that our case study is focused on grape classification, the primary techniques for this task are outlined below.

\textbf{Processing of hyperspectral signature}. Remotely sensed data is subject to various factors, such as sensor-related random errors, atmospheric and surface effects, and acquisition conditions. Therefore, radiometric correction is performed to obtain accurate data from the Earth's surface. Although the literature in this field covers numerous topics, it primarily focuses on satellite imaging. While some of the techniques studied can be applied to UAV imaging, other topics are irrelevant to our case study. For instance, atmospheric effects like absorption are not significant in close-range work. However, due to low flight altitudes, UAV instability, and varying viewing angles, pre-processing operations can be challenging \citep{jakob_need_2017}.

Most studies that involve the classification of satellite images use standard datasets with radiometric corrections, provided by the Grupo de Inteligencia Computacional (GIC) \citep{m_grana_hyperspectral_nodate}. In the case of UAV hyperspectral imaging, various corrections are necessary to obtain precise data, including geometric and radiometric corrections and spectral calibrations \citep{adao_hyperspectral_2017}. Geometric distortions are primarily caused by UAV instability and the acquisition technique, with push-broom sensors showing higher geometric distortions that can be reduced using stabilizers. Geometric correction can be achieved through an Inertial Navigation System (INS), Global Positioning System (GPS), and Digital Elevation Model (DEM). Although commercial software is available for this approach, it requires a high-precision system for accurate correction. Alternatively, Ground Control Points (GCPs) have been extensively utilized to ensure correct positioning \citep{ramirez-paredes_low-altitude_2016}. In addition, dual acquisition of visible and hyperspectral imagery enables matching both data sources \citep{jurado_efficient_2021, xue_compact_2021, ramirez-paredes_low-altitude_2016}, with visible data being more geometrically accurate. Another technique that has been shown to assist with geometric correction is feature matching among overlapping images \citep{akhoundi_khezrabad_new_2022}.

In a similar way to geometric distortions, radiometric anomalies can also be fixed with software tools provided by the hyperspectral manufacturer. The aim is to convert the Digital Numbers (DN) of the sensor to radiance and reflectance of Earth's surfaces, regardless of acquisition conditions. Therefore, the latter result must be applied to Deep Learning techniques for their implementation over any hyperspectral dataset. The coefficients required for this correction are generally calibrated in the laboratory, but they may vary over time \citep{adao_hyperspectral_2017}, which may affect the radiometric correction. Grayscale tarps, whose reflectance is known, can be used to support this process and perform linear interpolations to calibrate the acquired DN \citep{lucieer_hyperuasimaging_2014} using the empirical line method (ELM) \citep{aasen_quantitative_2018, sousa_uav-based_2022}. To perform the linear interpolation for the radiometric correction, it is necessary to have dark and grey/white references, which are usually obtained from isotropic materials that have a grayscale palette and exhibit near-Lambertian behaviour \citep{jakob_need_2017, sagan_data-driven_2022, duan_land_2013}. An alternative approach is to acquire radiance samples, which can be used with fitting methods such as the least-square method (LSM) to transform DNs. However, factors such as the vignetting effect, which causes a drop in intensity, can result in a different radiometric response in pixels of the same bands \citep{yang_dom_2017}.

There have been studies on atmospheric corrections for UAV flights, which involve correlating the at-sensor radiance to the surface hemispherical-directional reflectance function (HDRF) measured in the laboratory beforehand. Grayscale palettes in Lambertian materials can be used for this purpose, as documented in previous works \citep{lucieer_hyperuasimaging_2014}. While physically-based methods have also been explored, they tend to be more time-consuming. In addition, public repositories such as the Aviris data portal (California Institute of Technology) \citep{california_institute_of_technology_aviris_nodate} offer atmospherically corrected datasets. Other widespread hyperspectral processing software, such as ENVI and Headwall SpectralView, include semi-automatic correction algorithms \citep{queally_flexbrdf_2022, jia_kernel-driven_2020, sagan_data-driven_2022}.


\textbf{Traditional hyperspectral classification}. Deep Learning methods have recently become the preferred approach for classifying hyperspectral imagery. However, earlier techniques relied on comparing the acquired data to reference reflectance shapes that were ideally measured in a laboratory. The primary objective of these methods was to measure the similarity between labelled and unlabelled spectral shapes. Spectral libraries, containing data measured from a spectrometer, were used for this purpose. For instance, \cite{kokaly_usgs_2017}, \cite{dutta_characterizing_2017}, and \cite{matusik_data-driven_2003} provided spectral libraries for minerals, trees, and daily surfaces, respectively. These methods varied from the widely used Euclidean distance to more sophisticated techniques such as Spectral Angle Matching (SAM), Cross-Correlogram Spectral Matching (CCSM), and probabilistic approaches like Spectral Information Divergence (SID) \citep{pu_hyperspectral_2017}. Among these techniques, SID and CCSM have been found to perform better in mineral classification from Aviris data. Similarly, \cite{van_der_meer_effectiveness_2006} proved that SAM introduced much more confusion than SID and Spectral Correlation Matching (SCM) in a case study of mineral classification. SAM has the advantage of being invariant to different scales, making it useful for heterogeneous acquisition devices and conditions. Other techniques derived from SAM include Spectral Correlation Angle (SCA), based on the Pearson correlation coefficient, and Spectral Gradient Angle (SGA) \citep{ren_novel_2022}. In addition to similarity, angular and probabilistic measures, the literature describes error and colourimetric methods. The former group includes the widely used Mean Square Error (MSE), Root Mean Square Error (RMSE), Mean Relative Absolute Error (MRAE), Back-Projection MRAE (BPMRAE) and the Peak Signal-to-Noise Ratio (PSNR) \citep{agarla_analysis_2021}. More recently, \cite{kumar_new_2021} introduced three new metrics (Dice Spectral Similarity Coefficient (DSSC), Kumar–Johnson Spectral Similarity Coefficient (KJSSC), and a hybrid of the previous, KJDSSC$_{\textit{tan}}$, that outperformed traditional techniques on mineral and vegetation classification.

Colourimetric measures involve measuring distance in various colour spaces, such as pro-Lab. \cite{agarla_analysis_2021} have compared all these techniques to assess their correlation and determine the most significant ones. Additionally, spectral derivatives, which are finitely approximated considering the spectral signatures and wavelength distance, are used to remove or compress illumination variations resulting from acquisition conditions \citep{fernandes_grapevine_2019, pu_hyperspectral_2017}. These techniques are still applied when the reflectance profile of different materials exhibits notable variations. Furthermore, semi-automatic classification methods have been reviewed for situations where classification involves only a few labels. For instance, \cite{ahmed_applied_2021} used Principal Component Analysis (PCA) to extract features from multispectral imagery and proposed a vegetation index to label different trees. \cite{padua_monitoring_2020} described similar work on classifying chestnuts with phytosanitary problems using two new indices: ExNIR and ExRE, where Ex refers to Excess. However, these techniques are not suitable for differentiating a significant number of vineyard varieties since they all exhibit similar shapes. We refer the reader to \cite{shanmugam_spectral_2014} for an in-depth revision of spectral matching and attributes conditioning the construction of spectral libraries.

\textbf{Hyperspectral transformation and feature extraction}. In this section, we discuss the transformations that facilitate classification using traditional methods. Due to the extensive coverage of land by satellite imagery, it is uncommon for hyperspectral pixels to depict the spectral signature of a single material. Therefore, there is a prevalent topic in the hyperspectral literature, which involves breaking down the acquired Earth's surfaces by analyzing the hyperspectral images. The problem is illustrated with $\rho = \textit{MF} + \epsilon$, where $M$ is the spectral signature of different materials, $F$ is the weight, $\epsilon$ is an additive noise vector and $\rho$ is an $L \times 1$ matrix where $L$ is the number of bands. Hence, the difficulty of finding a solution to $M$ and $F$ is lowered if $M$ is fixed, i.e., the end-member signatures are known. The Multiple end-member spectral mixture analysis (MESMA) was the initial approach taken, followed by the Mixture-Tuned Matching Filtering Technique (MTMF), which eliminates the need to know end-members in advance. This approach was further refined with the Constrained Energy Minimization (CEM) method, which effectively suppresses undesired background signatures.

The current state-of-the-art techniques for Linear Mixture Models (LMM) can be categorized based on their dependency on libraries. Additionally, the level of supervision and computational cost also determines the classification of these techniques. The taxonomy of methods, as described by \cite{borsoi_spectral_2021}, varies depending on these factors. For instance, Bayesian methods and Local unmixing do not require to known end-member signatures, although Bayesian-inspired approaches are less supervised and more time-intensive. Besides MESMA, other proposed methods that require spectral signatures are based on Artificial Intelligence techniques such as Machine Learning and Fuzzy unmixing. The latter is less supervised but more time-consuming. In recent years, interest in Deep Learning (DL) has grown, with techniques such as autoencoders, Convolutional Neural Networks (CNN), and Generative Adversarial Networks (GAN) being utilized for training with synthetic data \citep{bhatt_deep_2020}. Non-negative matrix factorization (NMF) has also attracted attention as it can extract sparse and interpretable features \citep{hruska_machine_2018}. Recently, the incorporation of spatial information into hyperspectral unmixing has been investigated \citep{shi_incorporating_2014}. This involves considering the surrounding pixels using kernels of varying sizes and shapes, such as squared, cross, or adaptive. Weights can also be assigned based on the distance to the centre and the measured similarity using functions like SID, SAM, Euclidean distance, etc. Current state-of-the-art methods, such as NMF, have been combined with spectral information \citep{zhang_spectral-spatial_2022}.

Besides discerning materials, the results of Hyperspectral Imaging (HSI) present a large number of layers that can be either narrowed or transformed, as many of them present a high correlation. Otherwise, the large dimensionality of HSI data leads neural networks and other classification algorithms to be hugely complex. Accordingly, the most frequent projection method is PCA \citep{jiang_rapid_2022, shenming_new_2022, lu_hyperspectral_2022}, which projects an HSI cube of size $X \times Y \times \lambda$ into $D \times B$, where $D$ has a size of $X \times Y \times F$, and $B$ is a matrix such as $F \times \lambda$. In this formulation, $F$ is the number of target features \citep{amigo_hyperspectral_2019}. Independent Component Analysis (ICA) is a variation of PCA that not only decorrelates data but also identifies normalized basis vectors that are statistically independent \citep{pu_hyperspectral_2017}. Least Discriminant Analysis is another commonly used technique, but it is primarily applied after PCA to increase inter-class and intra-class distance \citep{shenming_new_2022}. In the literature, it is also referred to as Partial Least-Square Discriminant Analysis (PLS-DA), mainly as a classifier rather than a feature selection method.

Instead of projecting features into another space, these can be narrowed into the subset with maximum variance according to the classification labels of HSI samples. There are many techniques in this field, including the Successive Projection Algorithm (SPA), which reduces colinearity in the feature vector. The Competitive Adaptive Reweighted Sampling (CARS) method selects features with Monte-Carlo sampling and iteratively removes those with small absolute regression coefficients. Two-Dimensional Correlation Spectroscopy (2DCS) aims to characterize the similarity of variance in reflectance intensity. \cite{liu_dimension_2019} used the Ruck sensitivity analysis to discard bands with a value below a certain threshold. \cite{agilandeeswari_crop_2022} calculated the band entropy, vegetation index, and water index for wavelength subsets, generating a narrower cube only with bands above three different thresholds. Finally, the work of \cite{santos-rufo_wavelength_2020} presents an in-depth evaluation of methods based on Partial Least Squares (PLS) regression. To this end, HSI data from olive orchards were first narrowed and then classified with LDA (Least Discriminant Analysis) and K-Nearest Neighbours (KNN). In conclusion, the Lasso method \citep{friedman_regularization_2010} as well as Genetic algorithms \citep{mehmood_review_2012} showed the best performance with LDA. 

Remote sensing data typically contains inherent noise, which means it is rarely used as-is. To address this issue, \cite{gutierrez_--go_2018} used a combination of Standard Normal Variate (SNV) and de-trending to remove the scatter effect. Then, they smoothed the hyperspectral signature by applying Savitzky-Golay filtering with different step sizes over the first and second derivatives.


\textbf{Classification of hyperspectral imaging with ML and DL}. This section focuses on reviewing studies related to the classification of vineyard varieties using HSI. Although there are numerous studies on HSI classification, only those relevant to vineyard varieties will be discussed. In addition, state-of-the-art DL networks achieving high accuracy in HSI classification will also be briefly reviewed.

Despite there exists considerable research on segmentation, only a few studies have addressed the classification of vineyard varieties using RGB and multispectral imagery. In these studies, binary masks or grayscale maps were first extracted to distinguish soil, shadows, and vineyards. Clustering, line detection, or machine learning (ML) algorithms and artificial neural networks (ANNs) were then applied to segment vineyard rows \citep{fuentes-penailillo_using_2018, karatzinis_towards_2020, hajjar_vine_2021, padua_monitoring_2020, padua_vineyard_2022, poblete-echeverria_detection_2017}. Geometrical information from depth maps, digital elevation models (DEMs), LiDAR data, and photogrammetric reconstructions were also assessed \citep{kerkech_vine_2020, aguiar_localization_2022, jurado_automatic_2020}. DL approaches for semantic segmentation and skeletonization algorithms have also been discussed \citep{kerkech_vine_2020-1, barros_multispectral_2022, nolan_automated_2015}. Further insight into this field is provided in \cite{li_performance_2020}. 

The classification of different vineyard varieties has been previously achieved with traditional methods and proximal hyperspectral sensing. In the work of \cite{gutierrez_--go_2018}, samples were selected by manually averaging the variety signature and filtering those with high correlation to such a signature. Support Vector Machine (SVM) and Multilayer Perceptron (MLP) were then trained with k-fold to distinguish thirty varieties (80 samples for each one), with the latter obtaining a recall close to one. \cite{kicherer_phenoliner_2017} also presented a land phenotyping platform that segments grapes from the depth map and discerns between sprayed and non-sprayed leaves. To this end, several learning models were tested: LDA, Partially Least Square (PLS), Radial Basis Function (RBF), MLP and soft-max output layer (PNET), with RBF and PLS showing the best results. Besides phenotyping, the following work is aimed at detecting diseases \citep{nguyen_early_2021, bendel_detection_2020, bendel_evaluating_2020} and plagues \citep{mendes_vineinspector_2022}. However, these applications formulate a binary problem where signatures of distinct classes are significantly different regarding scale \citep{bendel_detection_2020} and shape \citep{bendel_detection_2020}. Despite this, previous learning models are also implemented (MLP, RBF, PLS and LDA) and almost achieved the perfect discrimination performance \citep{bendel_evaluating_2020}. \cite{nguyen_early_2021} conducted a study similar to ours, where they attempted to differentiate healthy and infected leaves with a comparable spectral signature. However, their data was obtained from land, and they used the flattened layer of 2D and 3D convolutional networks as input for Random Forest (RF) and SVM algorithms. They found that combining PCA reduction (50 features) and RF resulted in the best performance (97\%), and RF improved SVM classification regardless of data reduction. Additionally, ML and DL techniques have been extensively applied to various crops, such as maize, sugarcane, rice, and bread wheat, using both satellite and proximal imaging. Transfer learning, attention-based, and residual models are commonly used in the literature \citep{zhang_classification_2022}. A lightweight CNN composed of several inception blocks was also developed to classify up to 15 plant species \citep{liu_plant_2022}. The authors compared their proposed CNN model to other commonly used models for classifying RGB images, including AlexNet, VGGNet, and GoogLeNet. They found that the best results were achieved using a combination of six RGB and Near Infrared features, with an accuracy of 94.7\%. The use of PCA with only six features achieved an accuracy of 88\%. \cite{nezami_tree_2020} also applied a 3D CNN to classify three tree species using both hyperspectral and visible imaging, as well as canopy height models as input, with an overall accuracy below 95\%.

\begin{figure*}[ht]
    \centering
    \includegraphics[width=\linewidth]{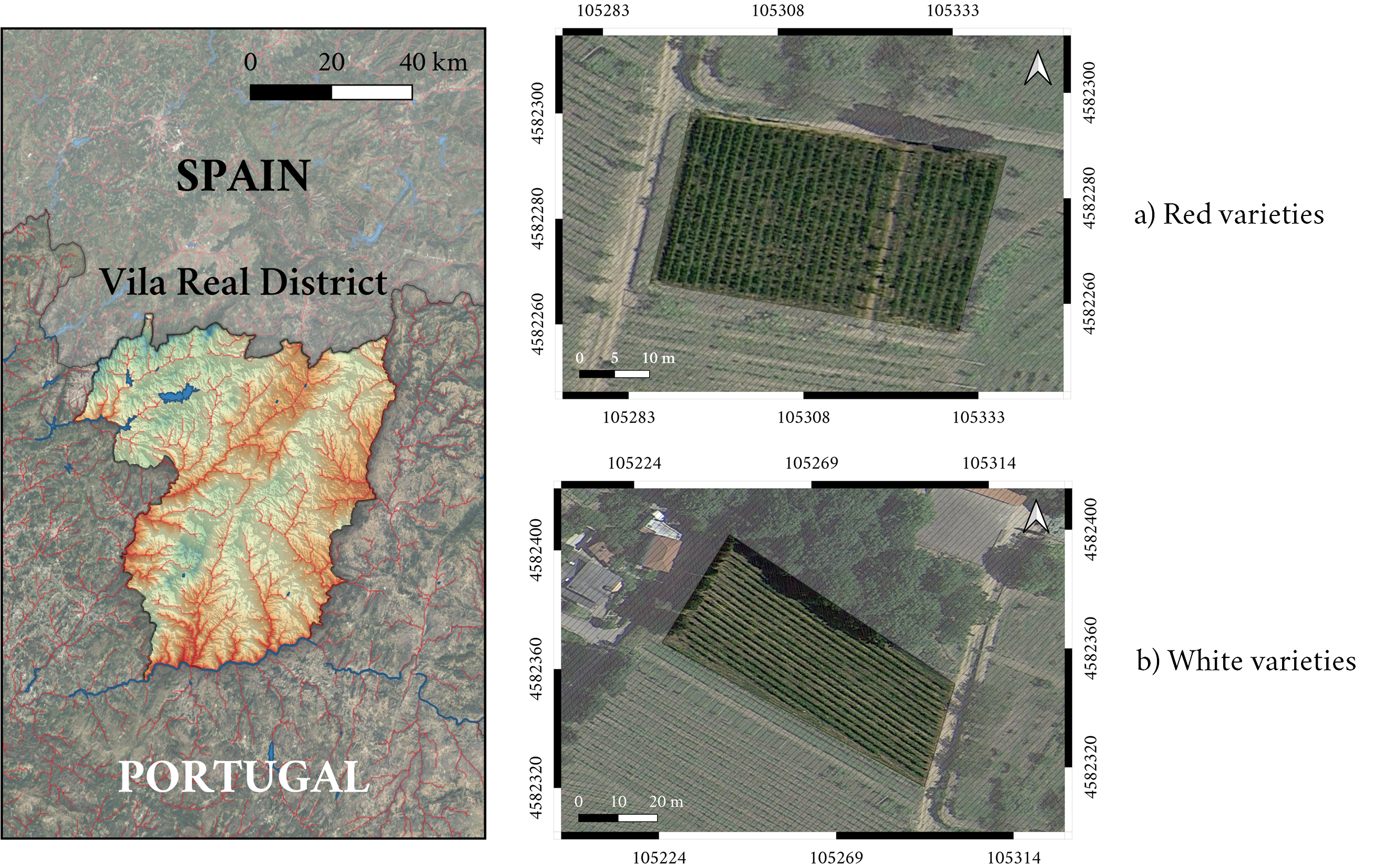}
	\caption{Overview of the areas surveyed with UAV hyperspectral imaging for the classification task. Two different vineyard crops are depicted according to their main variety: a) red and b) white. }
	\label{fig:study_area}
\end{figure*}

When it comes to DL for the classification of satellite hyperspectral imaging, it is more current than UAV imaging. A standard dataset is available for this purpose, on which several experiments have been conducted. Among them, the top-performing models based on overall accuracy (OA) are discussed below. \cite{zhong_whu-hi_2020} both published an HSI dataset and proposed a simple CNN with Conditional Random Field (CRF) to extract spatial relations among data, even with the presence of gaps. They obtained an OA of 98\% and 94\% over their own HSI dataset. \cite{moraga_jigsawhsi_2022} presented an Inception-based model with parallel convolutional pipelines of increasing size, achieving near-perfect classification. \cite{chakraborty_spectralnet_2021} proposed the SpectralNet model, which combines wavelet decompositions with a traditional convolutional path (OA: 98.59\%-100\%). \cite{roy_hybridsn_2020} developed HybridSN, which includes both spectral-spatial and spatial feature learning using 3D and 2D convolutional layers (OA: 99.63\%-100\%). \cite{roy_attention-based_2021} later introduced a network based on residual blocks and spectral-spatial attention modules with varying architecture (start, middle and ending ResBlock) (OA: 98.77\%-99.9\%). Lastly, the A-SOP network \citep{xue_attention-based_2021} proposed a module composed of matrix-wise operations that output a second-order pooling from the attention weights, after extracting the first-order features (OA: 98.68\%-100\%). 

Similar to Moraga and Duzgun's work in 2022, the FSKNet model also employs a combination of 2D and 3D convolutional layers with an intermediate separable convolution to reduce training latency while achieving comparable overall accuracy results. The FSKNet model achieves an OA above 99\% with significantly fewer parameters and a shorter training time. This work focuses on single output classification, where the output is either hot-encoded or given as a single value. However, other approaches have gained attention, such as contrastive learning and multi-instance segmentation, which propose outputs of higher dimensionality. \cite{zhu_spectral-spatial-dependent_2021} investigated pixel-wise classification of HSI patches using semantic segmentation, using the popular U-Net architecture with additional convolutional LSTM and attention-based mechanisms. \cite{xin_convolution_2022} used transformers to independently encode spatial and spectral features and then combine them. Contrastive learning has also been used to address the lack of labelled datasets, where HSI patches and 1D data are jointly used during training so that the network learns by comparing pairs of samples \citep{guan_spatial-spectral_2022}. Finally, \cite{meerdink_multitarget_2022} accurately labelled HSI with multi-instance learning by creating bags of samples with distinct labels.

\section{Materials and methods}

The structure of this section is as follows: firstly, a brief explanation of the study area and sensors is provided. Next, the challenges of classifying vine varieties are introduced by the collected data. Subsequently, UAV imagery is utilized to differentiate between phenotypes of white and red root variants. To achieve this, a CNN architecture is proposed, which is evaluated against previously reviewed work with impressive OA results.

\begin{figure*}[ht]
    \centering
    \includegraphics[width=\linewidth]{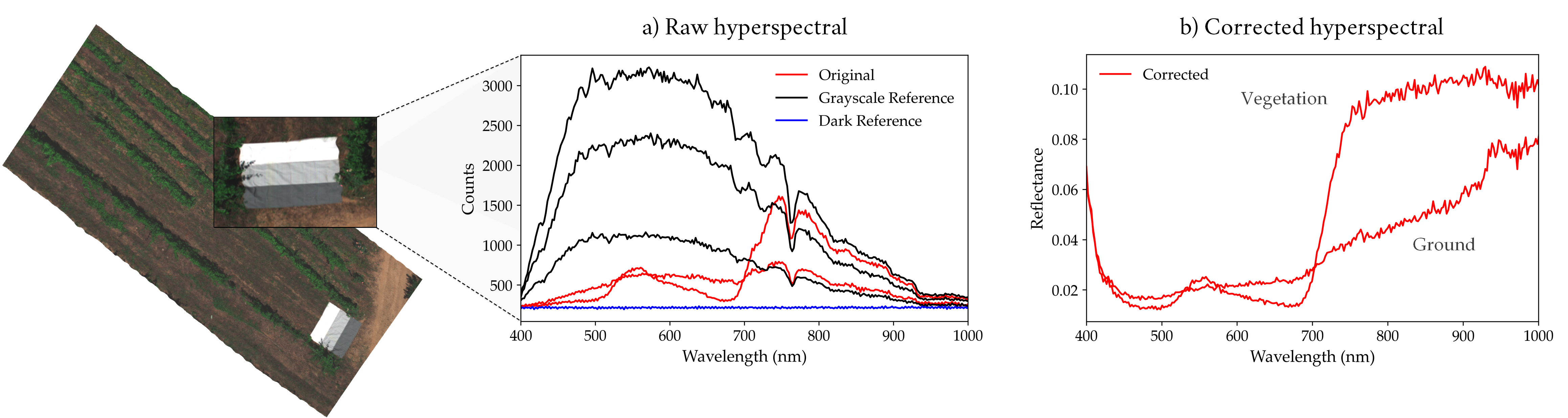}
	\caption{Conversion of a) hyperspectral DNs into b) reflectance using a white and dark reference. The three grey levels are sampled in a). }
	\label{fig:spectral_view}
\end{figure*}

\subsection{Study area}

The vineyards used as study areas in this work are situated in the Northern region of Portugal, specifically in Vila Real (Figure \ref{fig:study_area}). Each vineyard plot is dedicated to either red or white grapevine variants, and each grapevine variety is cultivated in one or more contiguous rows. Data acquisition was conducted with two different UAV flights to collect HSI swaths. 

\newcommand{\varietySpacing}{\hskip 0.3in}
\renewcommand{\arraystretch}{1.1}
\begin{table}[bp]
\centering
\caption{Summary of acquired information regarding different grapevine varieties. For each variety, the number of field rows and samples obtained by labelling UAV imagery is shown.\\ }
\label{table:grape_samples}
\begin{tabular}{|l|l@{\varietySpacing}|l|l|}
\toprule
\textbf{Root} & \textbf{Variety} & \textbf{\#rows} & \textbf{\#samples}\\
\cmidrule{1-4}
\multirow{8}{*}{Red} & Alicante & 3 & 58,680\\
& Alvarhelao & 4 & 144,315\\
& Barroca & 3 & 35,656\\
& Sousao & 3 & 75,078\\
& Touriga Femea & 3 & 36,114\\
& Touriga Francesa & 3 & 71,547\\
& Touriga National & 3 & 53,620\\
& Tinta Roriz & 4 & 67,157\\
\cmidrule{1-4}
\multirow{9}{*}{White} & Arito Do Douro & 1 & 92,432\\
& Boal & 3 & 44,654\\
& Cercial & 1 & 105,384\\
& Codega Do Ladinho & 3 & 261,228\\
& Donzelinho Branco & 1 & 98,304\\
& Malvasia Fina & 3 & 242,412\\
& Moscatel Galego & 1 & 101,885\\
& Nascatel Galego Roxo & 1 & 92,432\\
& Samarrinho & 1 & 77,229\\
\cmidrule{1-4}
\multicolumn{3}{|r|}{Total} & 1,658,127\\
\bottomrule
\end{tabular}
\end{table}
\renewcommand{\arraystretch}{1}

\subsection{Material}

The DJI Matrice 600 Pro (M600) hexacopter equipped with a Nano-Hyperspec from Headwall was used for the UAV flight. The Ronin-MX gimbal was employed to minimize geometric distortions in HSI acquisition. The lens has a focal length of 12 \si{\milli\meter}, covering 21.1\textdegree. The HSI swaths have 270 spectral bands and a width of 640, with the height depending on the flight plan. The spectral range goes from 400 \si{\nano\meter} to 1,000 \si{\nano\meter}, with a uniform sampling of 2.2 \si{\nano\meter} that increased to 6 \si{\nano\meter} at half maximum. The UAV's location was captured at different timestamps using two positioning antennas, and angular data were recorded using an Inertial Measurement System (IMU). The flight was planned using Universal Ground Control Station at an altitude of 50 \si{\meter} with a 40\% side overlap. The red and white varieties were surveyed with 8 and 5 swaths, respectively. Table \ref{table:grape_samples} provides a summary of the number of samples concerning each grape variety.

\subsection{Pre-processing of hyperspectral data}
\label{sec:preprocessing}

This section briefly describes the process of obtaining reflectance data from raw hyperspectral imagery, which is illustrated in Figure \ref{fig:spectral_view}. The hyperspectral data was collected using a drone and processed using Headwall SpectralView\texttrademark \hspace{1mm} software. Several swaths were captured for each study area, and a white sample was marked from the white area in a grayscale tarp, while a dark reference was obtained by collecting a hyperspectral sample with the lens cap on before the flight. The sensor exposure and frame period were adjusted before the flight by pointing at a bright reference to avoid clamping samples from white surfaces. The white and dark references were then used to convert the raw data to reflectance. The ortho-rectified swath in Figure \ref{fig:spectral_view} was obtained using high-resolution DEMs (25\si{\meter}) from Copernicus' observation program \citep{european_environment_agency_eu_2017} and the drone's GPS and IMU data. However, non-ortho-rectified swaths were used for the analyses presented in this paper to work with smaller image sizes and avoid distorting the hyperspectral signatures.

\subsection{Transformation of hyperspectral data}
\label{sec:hsi_transformation}

\begin{figure}[ht]
    \centering
    \includegraphics[width=\linewidth]{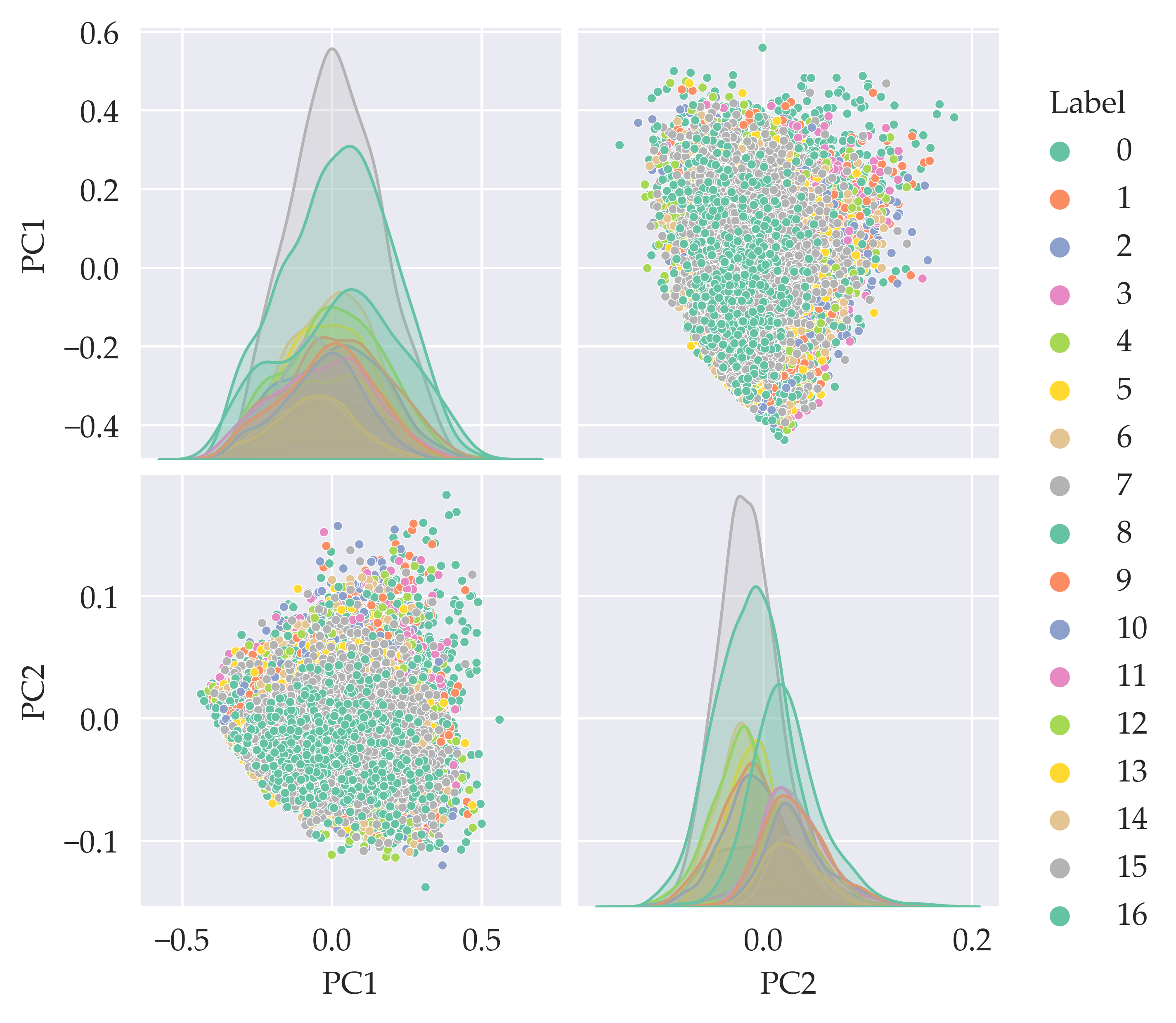}
	\caption{Distribution of hyperspectral samples according to the two components extracted using PCA.}
	\label{fig:pca_pairplot}
\end{figure}

\begin{figure}[ht]
    \centering
    \includegraphics[width=\linewidth]{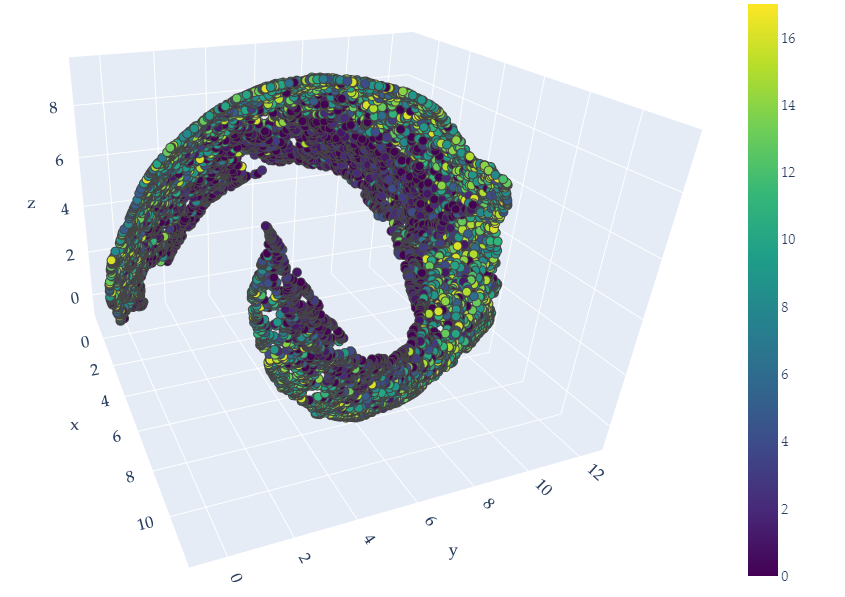}
	\caption{3D distribution of hyperspectral samples obtained by narrowing 50 components calculated with PCA using uMAP. }
	\label{fig:pca}
\end{figure}

The analysis of the corrected reflectance data involved several steps to observe and differentiate the spectral signatures from different varieties. Initially, PCA was employed to assess the clustering of each variety using the first two principal components. The distribution of the transformed signatures can be visualized in Figure \ref{fig:pca_pairplot}, without clear distinctions between different labels. Additionally, to further explore the clustering patterns, 50 features were extracted with PCA and following narrowed to three components using uMAP (Uniform Manifold Approximation and Projection for Dimension Reduction) \citep{mcinnes_umap_2020} (Figure Figure \ref{fig:pca}). Similar conclusions are drawn from the last investigation, showing no clear distinction among varieties.

\begin{figure}[ht]
    \centering
    \includegraphics[width=\linewidth]{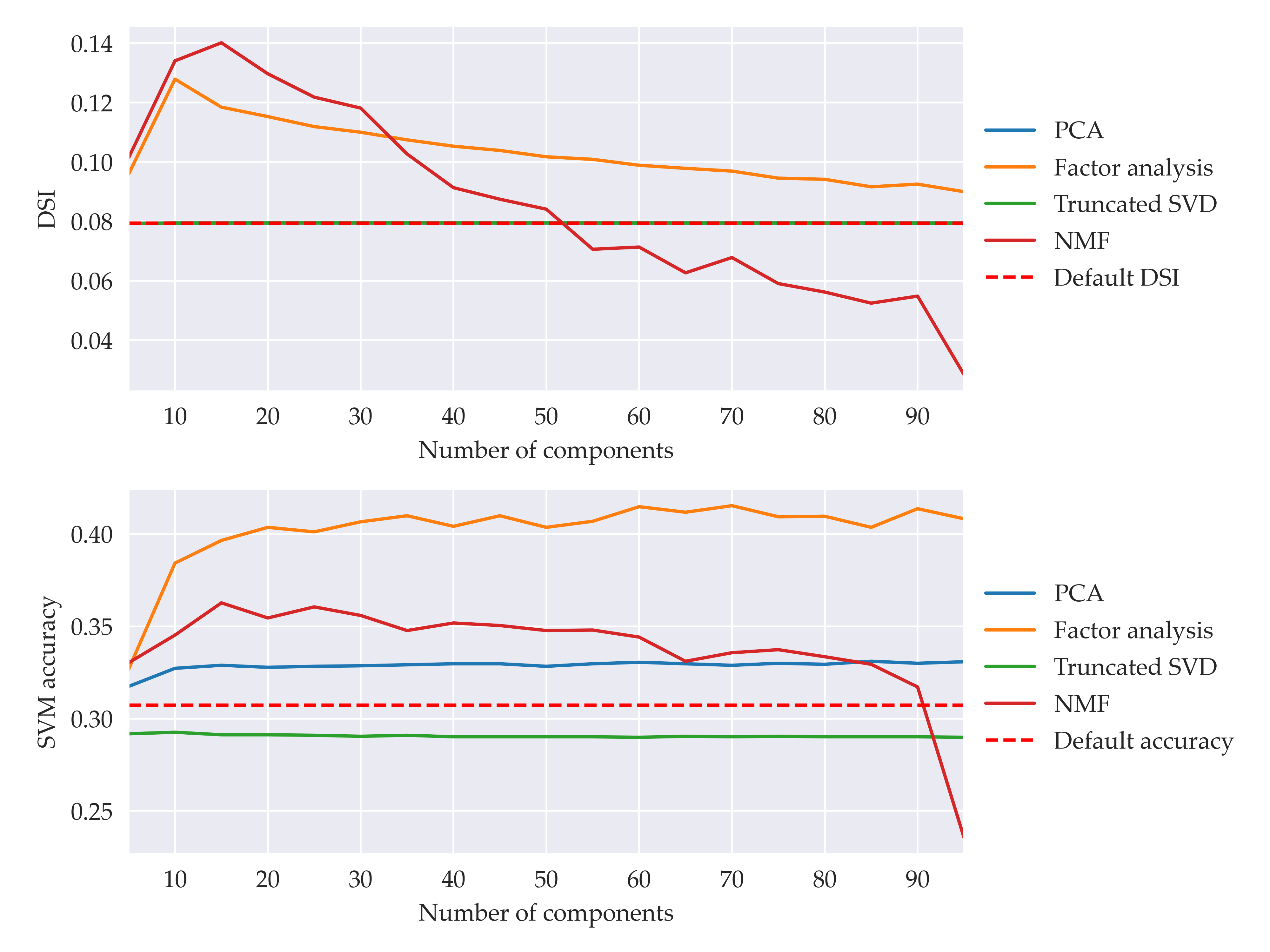}
	\caption{Results of experiments conducted to compare different feature transformation algorithms with different numbers of components. PCA, FA, NMF, and LSA (Truncated SVD) are evaluated using the DSI metric and the OA obtained by training an SVM model. The default DSI and accuracy are obtained from the original data with 140 features. }
	\label{fig:feature_reduction}
\end{figure}

To determine the most suitable feature transformation algorithm for the collected HSI data, we evaluated four algorithms: NMF, PCA, FA, and LSA. These were selected since they do not require the sample labels for their execution and therefore can work over yet not observed data. Additionally, LDA was included in these tests, despite requiring the labels. Remark that the starting data is composed of 140 bands after discarding the first and last layers, which are typically noisier. For each algorithm and varying numbers of features, from 5 to 95, two tests were performed. First, the Distance-based Separability Measure (DSI) was computed using the transformed manifold \citep{guan_internal_2020}. Subsequently, an SVM model was trained to predict the labels of the samples. Through this evaluation, it was determined that FA outperformed the other algorithms in terms of both metrics, especially with more than thirty-five features. The results of these experiments are summarized in Figure \ref{fig:feature_reduction}, which supports the use of FA with forty features in the following sections.

Furthermore, feature transformation algorithms, such as FA, help to perform material unmixing to make the processing more robust against different background surfaces, including soil, low-vegetation and other man-made structures. Pixels from UAV-based hyperspectral swaths depict more than one material. Therefore, the classification of these data ought to work under different kinds of surfaces. As proposed in recent work, the material unmixing could be performed with NMF. To this end, hyperspectral swaths in reflectance units could be flattened to 1D ($n \gets h \cdot w$) with a dimensionality of $n \times m$, where $m$ is the number of features. Then, this flattened vector could be transformed into weight ($W_{n \times c}$) and component matrices ($C_{c \times m}$), where $c$ is the number of target surfaces (end-members). However, the number of materials visible in a single image (or vineyard varieties) is not known in nature. Therefore, material unmixing in nature is rather suited for the classification of significantly different signatures rather than performing fine-grained classification. On the other hand, the feature space can be transformed and narrowed to a few more representative features, instead of unmixing materials. In this regard, FA is also aimed at decomposing $A$ into $W \times C + \Theta$ without the non-negative restriction, with $\Theta$ being the measurement error \citep{bandalos_measurement_2018}. For this reason, FA is proven to be highly suitable to our case study, beyond outperforming the rest of the feature transformation methods in terms of separability and classification accuracy using an SVM model.  

\subsection{Automated training}
\label{sec:training}

The classification of vineyard varieties with UAV data can be hardly approached with 1D algorithms due to the high similarity of spectral signatures. This shortcoming was already observed in Figure \ref{fig:feature_reduction}, where SVM did not perform well for any number of features (OA always below 50\%). In this section, a method based on deep learning is described to classify 3D hyperspectral patches. Through this section, the proposed method is tuned to achieve a high generalization performance. 

\begin{figure*}[ht]
    \centering
    \includegraphics[width=\linewidth]{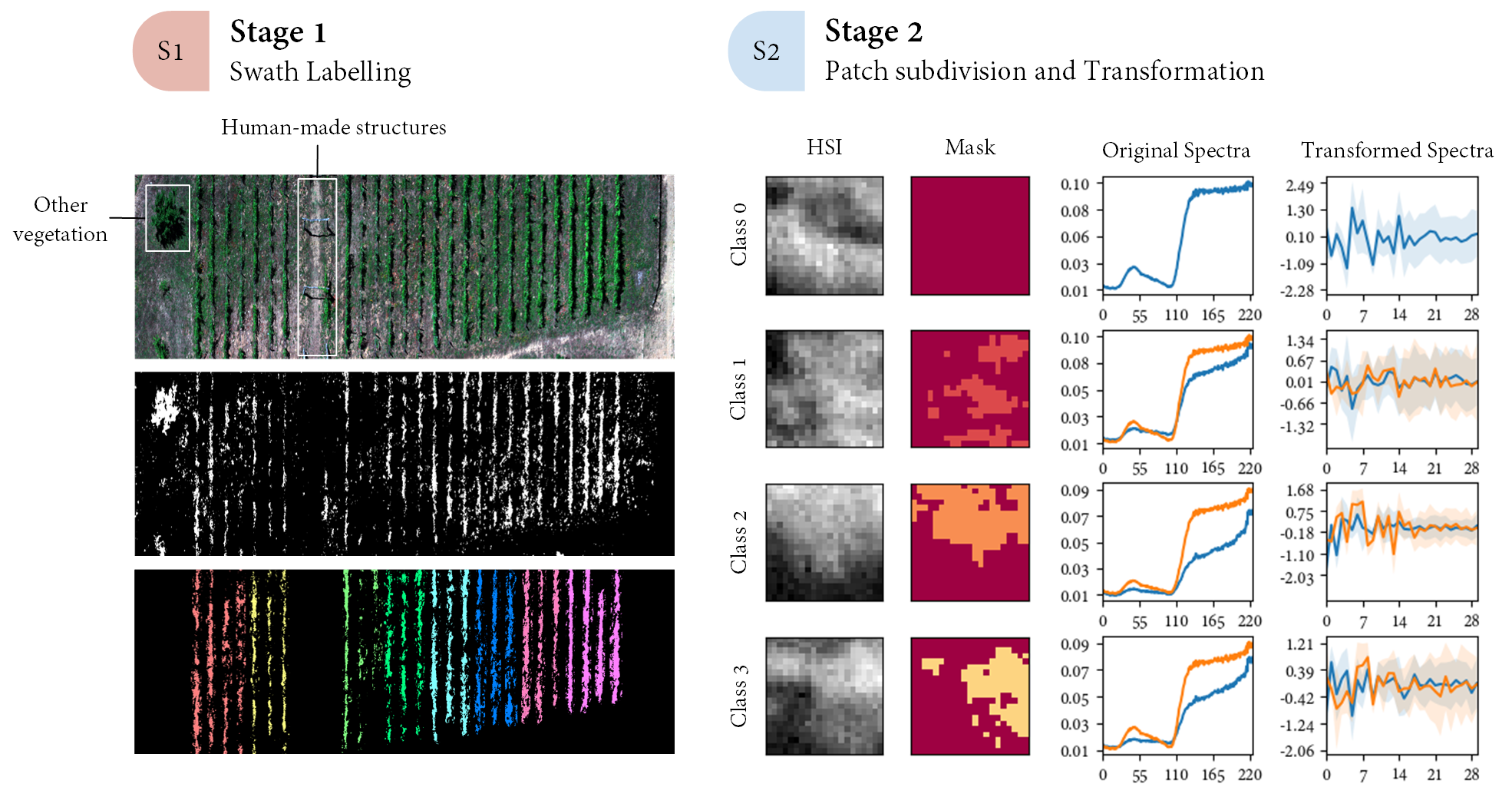}
	\caption{Overview of dataset preparation. First, a binary mask was generated using the NDVI and rows were organized into different groups to distinguish vineyard classes. Once pixels are processed as described in \nameref{sec:preprocessing}, both reflectance and labels were split into patches. The signatures on the right side show the original and transformed reflectance, including the variance per feature. }
	\label{fig:dataset}
\end{figure*}

\subsubsection{Dataset}

\begin{figure}[ht]
    \centering
    \includegraphics[width=\linewidth]{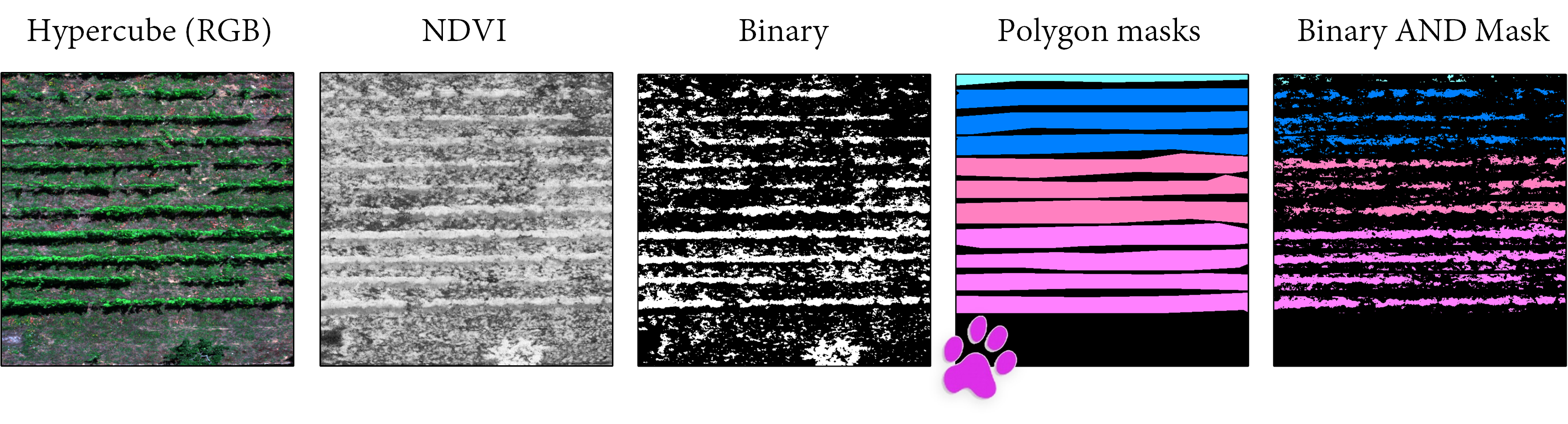}
	\caption{Workflow for manually labelling HSI swaths. First, the false RGB image is displayed. Then, the NDVI is extracted, followed by thresholding and marking with polygons using the Sensarea software. Finally, a Boolean operation, $\land$, is performed between the polygon and binary masks to obtain the final labelled regions.}
	\label{fig:manual_processing}
\end{figure}

Once radiometrically corrected, hyperspectral swaths were manually labelled as depicted in Figure \ref{fig:manual_processing} to distinguish different vineyard varieties. The Normalized Difference Vegetation Index (NDVI) was first extracted to differentiate vegetation from the ground, and images were then thresholded to create a binary mask from each swath. Following, these binary masks were annotated with Sensarea \citep{bertolino_sensarea_2012} by marking each row with a different polygon and colour, according to the variety. Some rows were marked with more than one polygon, in order to avoid annotating small vegetation clusters that do not belong to vineyards but to small vegetation. For this reason, different polygons were labelled with the same colours, also because some varieties were repeated in several rows. According to this, Table \ref{table:grape_samples} shows the number of collected samples for each variety and the number of cultivated rows. 

HSI cubes and masks were then split into 3D patches whose size is a matter of discussion in \nameref{sec:experimentation}. Dividing the hyperspectral swaths into patches for classifying pixels using their neighbourhood helps to partially suppress noise. Individual pixels are not substantial enough by themselves; instead, aggregations learned by kernels help to mitigate the noise. The window size used in this study is $23 \times 23$ by default, whereas previous work has used patches whose \textit{x} and \textit{y} dimensions range from 7 \citep{roy_attention-based_2021} to 64 \citep{chakraborty_spectralnet_2021}. Only \cite{liu_plant_2022} reported patches of much larger dimensionality ($200 \times 200$) for RGB images. The larger the patch, the deeper can be the network, though it also increases the number of trainable weights, the training time and the amount of data to be transferred into/from the GPU. Configurations using larger patch sizes are more suited to images with notable spatial features, such as close-range RGB imagery, whereas ours ought to be primarily discerned through spectral features. 

Instead of inputting the label of every patch's pixel, they were reduced to a single label corresponding to the centre of an odd-sized patch. Thus, the classification was performed per pixel rather than an overall semantic segmentation. Based on this, the hyperspectral samples were processed using the following steps: 1) separating the training and test samples at the outset, 2) fitting the FA and standardization only to the training samples to emulate a real-world application, and 3) transforming both the training and test samples using the fitted models (see Figure \ref{fig:dataset}). Standardization is utilized to eliminate the mean and scale reflectance to unit variance. By employing this approach, the CNN restricts the range of input HSI values, although the initial values are expected to differ due to various sensor exposures, frame periods, and environmental conditions across different flights. Regarding feature reduction, spectral bands were transformed and narrowed to $n \gets 40$ with FA. None of the fitted models requires the pixel's labels and thus are very convenient for their application in new and unlabelled areas. The resulting dataset is composed of 542,167 and 1,115,960 patches for red and white varieties, which were split into training (68\%), validation (12\%) and test (20\%) subsets. With this partitioning, a total of 368k samples are used for training on red varieties, whereas 758k are applied to white variety classification.

\subsubsection{Implementation}

In order to make this paper self-contained, a brief introduction about DL is detailed in this section. Firstly, Deep Learning refers to layered representations that evolve in a learning process where they are exposed to input data. Typically, the depth of these models is large enough to automatically transform data and learn meaningful representations of data. Despite these models being able to work over any kind of structured data, even 1D, it is here proposed that one's pixel neighbourhood may help with the phenotyping problem. Convolutional Neural Networks (CNN) have achieved remarkably good results in the computer vision field. Convolutions are designed to learn local features, rather than global, by applying element-wise transformations while sliding over input data. These are defined as rank-3 tensors defined by width, height and depth. The width and height determine how large is the neighbourhood of every element, whereas the depth is the number of different learned filters. Hence, a single filter is applied element-wise to compose a response map from input data, whereas the whole filter stack is known as a feature map. If several convolution operations are concatenated, these evolve from learning low-level details (e.g., edges) to high-level concepts. Since individual filters are applied element-wise, the learnt patterns are invariant to the position within the image \citep{chollet_deep_2021}. However, kernels may not be applied for every element. Instead, information can be compressed using steps greater than one, also known as the stride value. Another key concept in CNN is that not every node is connected, thus partially tackling the overfitting problem. Training and test errors ought to remain similar during training, which implies that the network is not learning the training data (overfitting) or generalizing excessively (underfitting). To avoid both situations, the capacity of the model must be tuned in terms of complexity to generalize and reach low training error.  

Trainable CNN layers are typically defined by a matrix of weights and biases applied over input data, $f(x; w, b)$, with $f$ being an activation function that allows solving non-linear problems. In this work, ReLU and Leaky ReLU have been applied to tackle the vanishing gradient problem, together with batch standardization. The latter operations work similarly to the standardizer applied as a preprocessing stage. On the other hand, $w$ and $b$ are updated during training with the objective of minimizing a loss function comparing ground truth and predicted values for supervised classification. This is done by an optimizer that changes these trainable parameters using the error gradient scaled by the learning rate, $\eta$. The greater $\eta$, the faster is achieved the convergence, though it can also lead to significant oscillations. This process is known as Gradient Descent (GD) \citep{kattenborn_review_2021}. As faster convergence may be necessary at the beginning, the decay and momentum concepts were introduced to downscale $\eta$ during training, thereby omitting abrupt changes.

Besides convolutions and normalization, there exist other layers to narrow data, avoid overfitting and output probabilistic values. The pooling operations, with max and average being the most popular, are aimed at downsampling input data. Dropout layers are used as a mechanism to introduce some noise into the training by zeroing out some output values, thus getting rid of happenstance patterns. Weight regularization also seeks to make the model simpler by forcing the weights to be small. Finally, the output units of the model are aimed at transforming features to complete the classification task. For a multi-label problem, the Softmax represents the probability distribution of a variable with $n$ possible values. 

The kind of problem and label representation is coupled with the cross-entropy function measuring the error. Sample labels were not hot-encoded to reduce storage footprint, and therefore, a sparse categorical cross-entropy as defined in Equation \ref{eq:crossentropy} is used for training in a multi-class problem. Otherwise, hot-encoding requires transforming labels into binary vectors of size $c$ that activate the indices of the sample label(s), with $c$ being the number of unique labels. 
\begin{align*}
    L_{\textit{CE}} = -\ln{\left(\hat{y}\left[y\right]\right)}
    \numberthis \label{eq:crossentropy}
\end{align*}
where $\hat{y}$ is the model's output as a vector of size $c$ with $\hat{y}\left[i\right], i \in \left[0, c - 1\right]$ indicating the probability of the sample to belong to the i-th class, and $y$ is the ground truth given by an integer value.

\subsubsection{Architecture and training}

\begin{figure*}[ht]
    \centering
    \includegraphics[width=.85\linewidth]{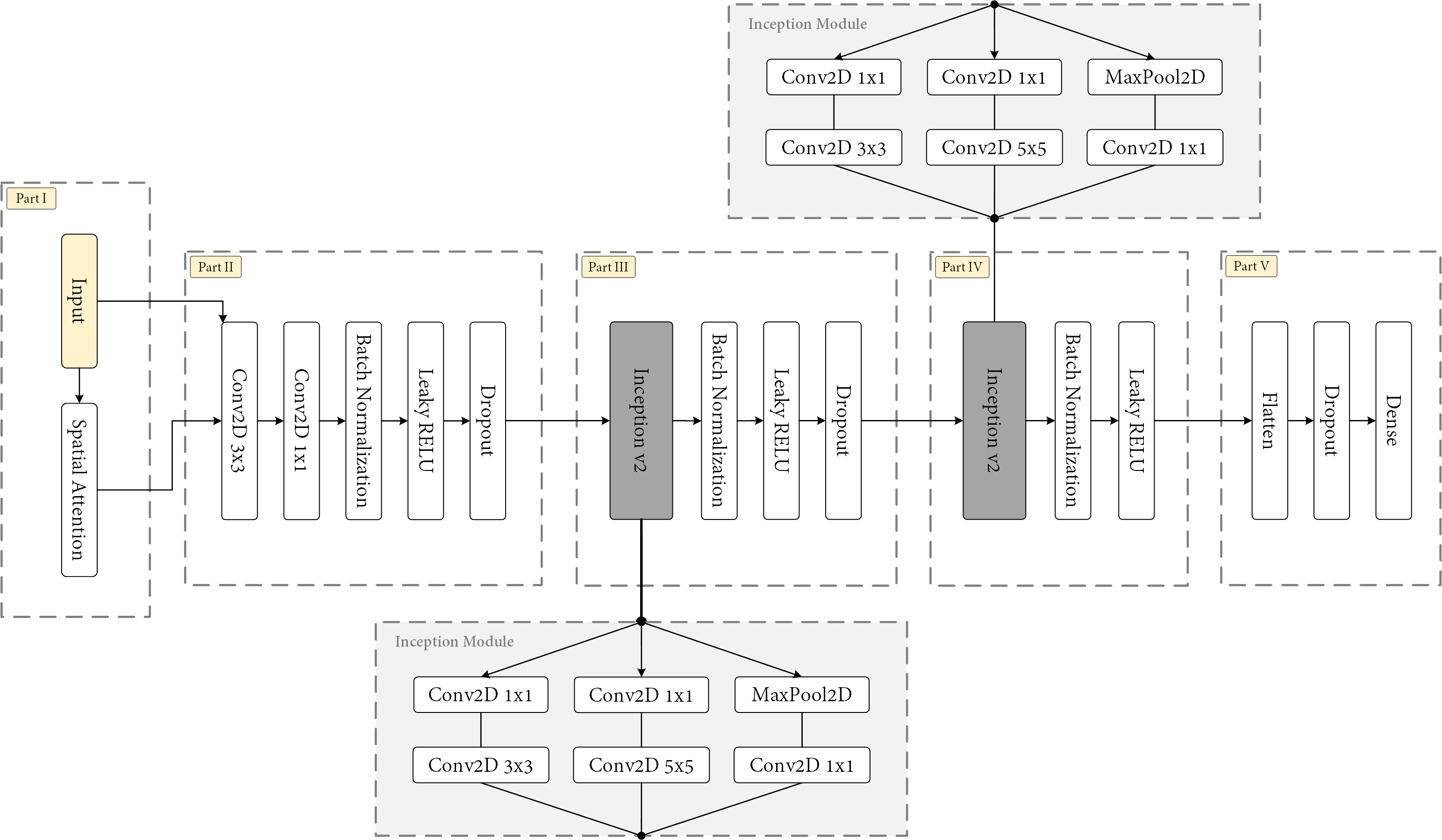}
	\caption{Scheme of the proposed CNN, highlighting four different parts as well as the structure of Inception blocks. }
	\label{fig:network}
\end{figure*}

Several architectures were checked over both datasets, transitioning from networks with a few layers to the network proposed in Figure \ref{fig:network}. Hyper-parameter tuning was also used to define the best values for dropout, activation and convolutional layers, including the number of filters, the percentage of zeroed weights, or the gradient of Leaky ReLU activation. Similarly, the final activation was also checked despite Softmax being usual for multi-class output. 

\newcommand{\kernelSize}[1]{#1 $\times$ #1}
\newcommand{\outputSize}[2]{#1 $\times$ #1 $\times$ #2}
\renewcommand{\arraystretch}{1.1}
\begin{table*}
\centering
\caption{Layer specifications of the proposed network. Inception blocks are simply named, but their layers were not expanded here to make this table more readable. \\ }
\label{table:cnn_architecture}
\begin{tabular*}{.9\textwidth}{LLLLL}
\toprule
Part & Layer & Kernel size & Strides & Output size\\
\midrule
\multirow{3}{*}{I} & Input & & & \outputSize{23}{40}\\
& Spatial Attention & & & \outputSize{23}{40}\\
& Concatenate & & & \outputSize{23}{80}\\
\midrule
\multirow{5}{*}{II} & Conv2D & \kernelSize{1} & 1 & \outputSize{23}{16}\\
& Conv2D & \kernelSize{3} & 2 & \outputSize{12}{16}\\
& Leaky ReLU ($\alpha \gets 0.1$) & & & \outputSize{12}{16}\\
& Batch normalization & & & \outputSize{12}{16}\\
& Dropout (0.2) & & & \outputSize{12}{16}\\
\midrule
\multirow{5}{*}{III} & Inception v2 & & 1 (Conv2D $1 \times 1$), 2 & \outputSize{6}{96} \\
& Batch normalization & & & \outputSize{6}{96}\\
& Leaky ReLU ($\alpha \gets 0.1$) & & & \outputSize{6}{96}\\
& Dropout (0.4) & & & \outputSize{6}{96}\\
\midrule
\multirow{5}{*}{IV} & Inception v2 & & 1 (Conv2D $1 \times 1$), 2 & \outputSize{3}{288}\\
& Batch normalization & & & \outputSize{3}{288}\\
& Leaky ReLU ($\alpha \gets 0.1$) & & & \outputSize{3}{288}\\
& Flatten & & & 2592\\
& Dropout (0.2) & & & 2592\\
\midrule
\multirow{1}{*}{V} & Softmax & & & 17\\
\midrule
\multicolumn{5}{c}{\#Trainable parameters: 562,227}\\
\multicolumn{5}{c}{\#Non-trainable parameters: 768}\\
\multicolumn{5}{c}{\#Parameters: 562,995}\\
\bottomrule
\end{tabular*}
\end{table*}
\renewcommand{\arraystretch}{1}

The input of the network is a single patch of size $23 \times 23 \times 40$. The spatial attention layer proposed by \cite{xue_attention-based_2021} is appended as an additional layer that is shown to help with the classification. The original paper proposed first and second-order pooling before and after the spatial attention module to transform the original features as we did with FA. 
Attention-based kernels have been recently used for HSI classification to provide more discriminative spectral-spatial features from input data. \cite{xue_attention-based_2021} applied attention-based as a link among the first data transformation, coined as a first-order feature operator, and a second-order kernel. \cite{roy_attention-based_2021} used an attention-based kernel in a similar fashion to \cite{xue_attention-based_2021}, though this one was replicated throughout the network pipeline. It was also part of a residual network where the input is simultaneously processed through 1) an attention kernel and 2) convolutional layers. The outputs of both pipelines are then concatenated at the ending point of each network's block. 

In this work, the attention-based kernel of \cite{xue_attention-based_2021} is applied over input data and the result is concatenated in $z$ with the original inputted data. The first operation in the attention-based thread is to normalize the data, from which the kernel will learn the weights. Such a kernel is a correlation matrix that learns the cosine distance between the central pixel and the neighbours. Then, learned weights are normalized through a softmax function that is shown to provide better convergence. With this in mind, the attention-based can be formulated as follows:
\newcommand{\inputMatrix}{{\mathcal{P}_{\textit{norm}}}^{M^2 \times B}}
\newcommand{\inputMatrixTranspose}{\left({\mathcal{P}_{\textit{norm}}}^{M^2 \times B}\right)^\intercal}
\begin{align*}
    \inputMatrix &\gets l_2 (\mathcal{P}^{M^2 \times B}) \numberthis \label{eq:spatial_attention_01}\\
    \mathcal{S}^{M^2 \times M^2} &\gets \left[\inputMatrix \inputMatrixTranspose\right] \numberthis \label{eq:spatial_attention_02}\\
    {\mathcal{S}_{central}}^{M^2 \times 1} &\gets \left[{{\mathcal{S}}_{\floor{\frac{M}{2}}}}^{M^2}\right]^\intercal \mathcal{K}^{M^2 \times 1}
    \numberthis \label{eq:spatial_attention_03}\\
    \mathcal{SA}^{M^2} &\gets \mathcal{S}^{M^2 \times M^2} \mathcal{K}^{M^2 \times 1} + \mathcal{B}^{M^2 \times 1} \numberthis \label{eq:spatial_attention_04}\\
    {\mathcal{P}_{\mathcal{SA}}}^{M^2 \times B} &\gets \textit{softmax}(\mathcal{SA}^{M^2}) \cdot \mathcal{P}^{M^2 \times B} \numberthis \label{eq:spatial_attention_05} 
\end{align*}
where $l_2$ refers to L2 normalization, $\mathcal{P}^{M^2 \times B}$ is a 3D patch resized from $\mathcal{P}^{M \times M \times B}$, $\mathcal{S}$, $\mathcal{S}_{\textit{central}}$ as well as $\mathcal{SA}$ are intermediate states and $\mathcal{P}_{\mathcal{SA}}$ is the result that is later concatenated with the original form.

Then, two similar blocks are included as Part II and Part III in Figure \ref{fig:network}. Both share the same structure: Inception block, normalization, activation and dropout. It is a frequent follow-up of convolutional layers \citep{li_faster_2022, xue_attention-based_2021}, with dropout being greater (0.4) for middle layers than the last and initial layers (0.2). Instead of using max-pooling to downsample the network, strides of size 2 in convolutional layers were observed to perform better. The network specifications are shown in Table \ref{table:cnn_architecture}. 

The Inception block was first proposed by \cite{szegedy_going_2014}. The first proposal consisted of a module with four parallel layers later concatenated: convolutional layers with different kernel sizes (1 for spectral features and 3 and 5 to obtain aggregations from surrounding pixels), and a max-pooling layer that works directly over input data. Accordingly, a response map with a large number of filters is obtained. The importance of each of them is determined by the following layers that will again downsample data. However, at the time this layout was considered to be prohibitive if the input layer has a large number of filters, especially for kernels of larger size. Therefore, $1\times1$ convolutions aimed at reducing data were attached before each one of the Inception threads (max-pool and convolutions with $\kappa > 1$). In this work, both proposals were used: the naïve is checked in the experimentation to increase the network's capacity, whereas the second is part of the proposed network. The latter compresses spatial data even more and is following connected to the network output. 

Finally, the model is fitted with training data and its performance is assessed with validation samples. For supervised problems like ours, data is composed of both samples and ground truth. In this work, the training samples were split into several sets according to the hardware limitations, and the model was iteratively trained during a significant number of iterations ($\varepsilon \gets 500$). Besides mitigating storage limitations, this leave-p-out cross-validation also helps to generalize by not training over the complete dataset. Furthermore, each one of these clusters is further split into small batches during a single iteration. The batch size must be large enough to include a balanced representation of samples. In this work, the batch size was set to $2^{10}$. This phase can be terminated early if no improvements are observed during $t \gets 20$ epochs. A summary of the hyperparameters used in this study can be found in Table \ref{table:hyperparameters}.

\renewcommand{\arraystretch}{1.1}
\begin{table}
\centering
\caption{Hyperparameters used during training.\\ }
\label{table:hyperparameters}
\begin{tabular}{|l@{\hskip 0.2in}|l@{\hskip 0.15in}|}
\toprule
\textbf{Hyperparameter} & \textbf{Value}\\
\cmidrule{1-2}
Patch size & 23\\
Patch overlapping & 22\\
Batch size & 1024\\
Epochs & 500\\
Learning rate & $1^{-5}$\\
Number of training splits & 9 \\
Transformations per split & 2 \\
Optimizer & RMS propagation \\
Loss function & Categorical crossentropy\\
Training split & 0.68\\
Validation split & 0.12\\
Test split & 0.2\\
\bottomrule
\end{tabular}
\end{table}
\renewcommand{\arraystretch}{1}

\begin{figure*}[ht]
    \centering
    \includegraphics[width=\linewidth]{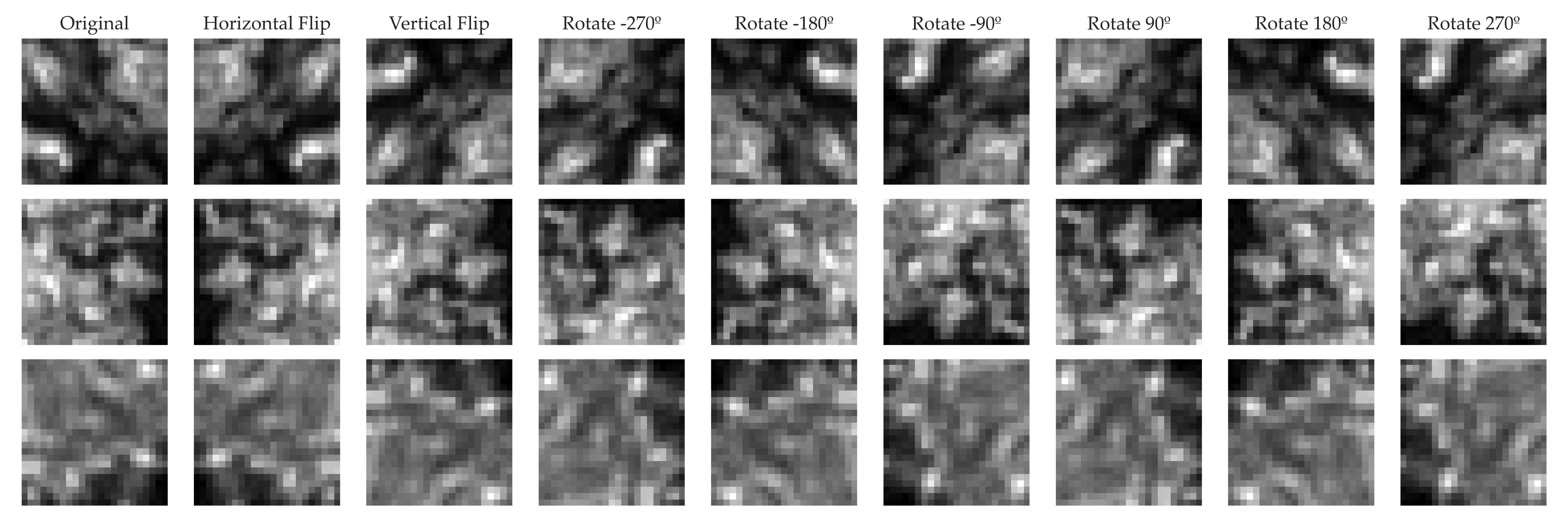}
	\caption{Transformations that can be probabilistically performed over every hyperspectral patch. }
	\label{fig:patch_transformations}
\end{figure*}

\subsubsection{Data sampling and regularization}
\label{sec:regularization}

\begin{figure}[ht]
    \centering
    \includegraphics[width=\linewidth]{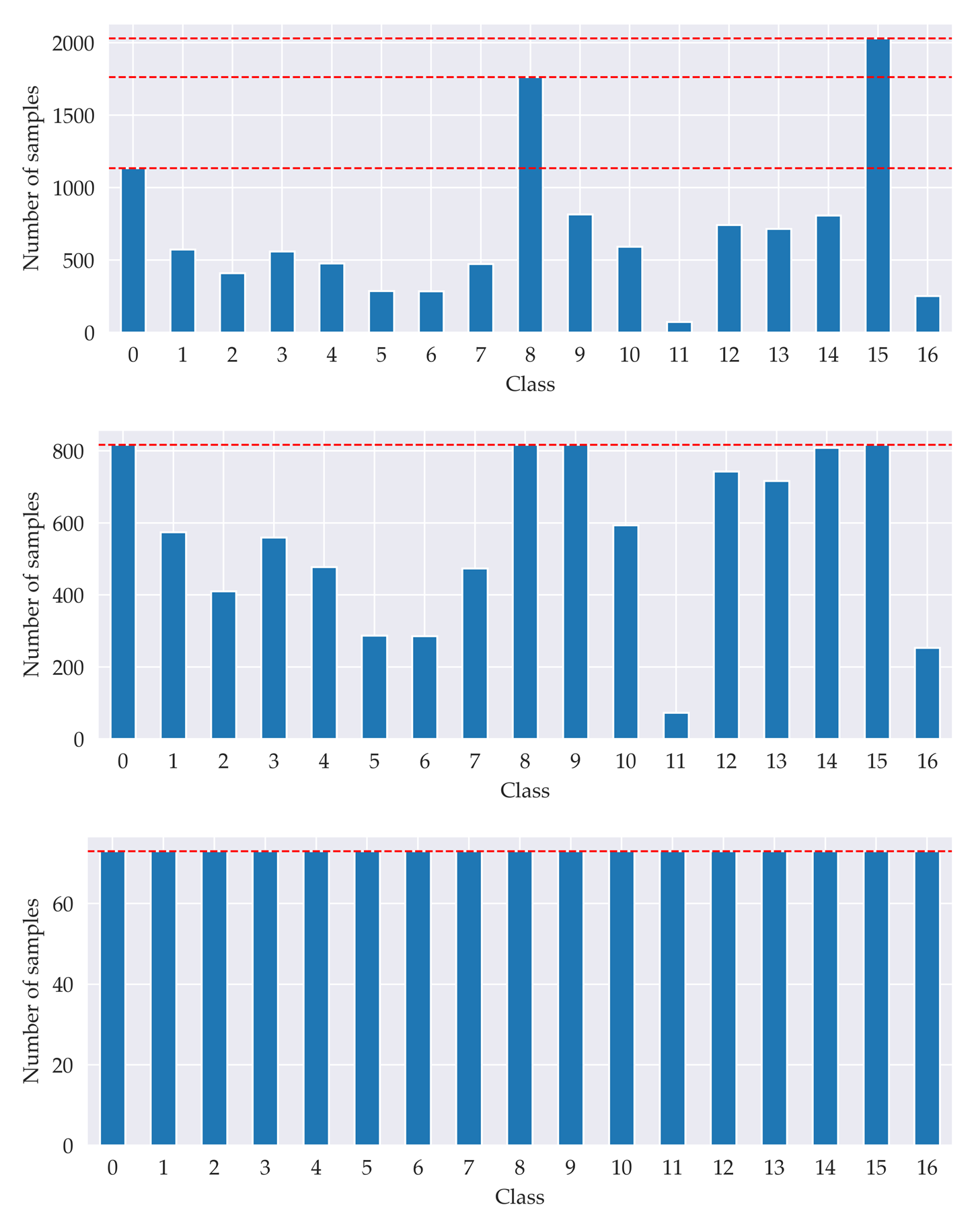}
	\caption{From top to bottom: initial distribution of samples per label, proposed narrowing, with only three groups being downsampled, and equalization of samples, leaving only seventy instances per group. }
	\label{fig:histogram_labels}
\end{figure}

It can be observed from Figure \ref{fig:histogram_labels} that the dataset is clearly not balanced. The number of vineyard rows differ in length and so does the number of examples for each variety. Instead of generating new feasible batches to upsample, a subset was obtained with different techniques. The objective is not to equalize the number of samples for every variety but rather to make it more balanced. Accordingly, the subsampling is performed by determining how many groups are downsampled; the larger it is, the more balanced gets the dataset at the expense of reducing the number of hyperspectral samples. In this regard, Figure \ref{fig:histogram_labels} compares the original distribution observed in a training batch, the utilized downsampling technique and a minority downsampling approach which leads to a huge decrease in usable patches. Besides balancing the dataset, which is split into several batches to make it fit in the GPU, the CNN is watched with a callback that saves the current best model and prevents saving an overfitted model.

Batches were probabilistically ($\mathcal{P} \gets 0.1$) transformed by performing rotations and orientation flips so that learn features are invariant to the flight's positioning conditions. Every possible transformation, regardless of probability, is shown in Figure \ref{fig:patch_transformations}. With this approach, each batch of the training dataset was processed twice; each one with a different random seed, and therefore, differently transformed patches. Hence, the regularization was controlled by the proposed downsampling and transformation sequences. Several considerations were also taken into account during the CNN design: 1) the CNN must not have a large number of trainable parameters to avoid overfitting and a sufficient number to cope with underfitting, and 2) dropout layers were included to randomly reset some output weights and thus lead to proper generalization.


\section{Experimentation and analysis}
\label{sec:experimentation}

\newcommand{\numberVariance}[2]{#1 $\pm$ #2}

\renewcommand{\arraystretch}{1.1}
\begin{table*}[hbt]
\centering
\caption{Overall results in terms of Overall Accuracy (OA), Average Accuracy (AA) and Kappa coefficient
with different methods. }
\label{table:overall_results}
\resizebox{\textwidth}{!}{\begin{tabularx}{\textwidth}{|X@{}|*7{c|}}
\toprule
Metric & Ours & LtCNN & Nezami et al. & JigsawHSI & SpectralNET & HybridSN & A-SPN\\
\cmidrule{1-8}
OA & \textbf{\numberVariance{98.78}{0.15}} & \numberVariance{74.33}{9.77} & \numberVariance{80.42}{0.59} & \numberVariance{73.89}{1.72} & \numberVariance{79.09}{0.55} & \numberVariance{63.46}{0.45} & \numberVariance{63.40}{0.69}\\
AA & \textbf{\numberVariance{98.94}{0.09}} & \numberVariance{73.09}{9.69} & \numberVariance{77.72}{0.43} & \numberVariance{73.55}{0.67} & \numberVariance{78.92}{0.29} & \numberVariance{63.04}{0.68} & \numberVariance{69.82}{0.49}\\
Kappa & \textbf{\numberVariance{99.67}{0.05}} & \numberVariance{91.15}{3.87} & \numberVariance{95.43}{0.28} & \numberVariance{90.27}{1.69} & \numberVariance{93.43}{0.15} & \numberVariance{89.68}{0.24} & \numberVariance{88.77}{0.31}\\
f1 & \textbf{\numberVariance{98.78}{0.15}} & \numberVariance{73.86}{10.29} & \numberVariance{80.38}{0.61} & \numberVariance{73.00}{2.41} & \numberVariance{79.05}{0.56} & \numberVariance{63.10}{0.70} & \numberVariance{61.69}{0.96}\\
\bottomrule
\end{tabularx}}
\end{table*}
\renewcommand{\arraystretch}{1}

The effectiveness of the proposed network is evaluated in this section. To this end, the classification experiments are jointly performed over various hyperspectral swaths of both surveyed areas. Results are presented in terms of overall accuracy (OA), average accuracy (AA), statistical kappa ($\kappa$) and f1-score. The first shows the percentage of correctly classified samples, the AA represents the average class-wise accuracy, the $\kappa$ coefficient is the degree of agreement between the classification results and the ground truth, and finally, the f1-score measures the model's precision by leveraging both precision and recall metrics. Several representative neural networks are compared with our method: LtCNN \citep{liu_plant_2022}, JigsawHSI \citep{moraga_jigsawhsi_2022}, SpectralNET \citep{chakraborty_spectralnet_2021}, HybridSN \citep{roy_hybridsn_2020}, Nezami et al. \citep{nezami_tree_2020} and A-SPN \citep{xue_attention-based_2021}. From these, only a few address airborne-sensing imagery \citep{liu_plant_2022}, whereas the rest are focused on satellite data. Hence, several considerations must be addressed: 1) some of these manuscripts apply different transformations to the input data and 2) the number of spectral bands also differ from our sensing tool. Therefore, the preprocessing pipeline was selected as the one providing better performance over our input data, either our pipeline or the one proposed in the reference work. However, FA showed a higher performance for every network if input data was transformed according to this fitting method, rather than the following: 
\begin{itemize}
    \item \cite{nezami_tree_2020}, \cite{liu_plant_2022} (LtCNN) used the corrected reflectance with no preprocessing. 
    \item \cite{moraga_jigsawhsi_2022} (JigsawHSI) used PCA, FA, SVD and NMF with 9-12 final features.
    \item \cite{liu_plant_2022} (A-SPN) and \cite{roy_hybridsn_2020} (HybridSN) used PCA to transform reflectance with $n \gets 15, 30$, respectively, whereas $n$ is unknown for \cite{xue_attention-based_2021}.
    \item \cite{chakraborty_spectralnet_2021} (SpectralNet) used FA with only 3 features.
\end{itemize}

Regarding implementation, all the tests were performed on a PC with AMD Ryzen Threadripper 3970X 3.6 GHz, 256 GB RAM, Nvidia RTX A6000 GPU and Windows 10 OS. The proposed CNN as well as the compared networks were implemented with Keras (version 2.10.0) and TensorFlow (version 2.10.1) in Python. CUDA 11.8 and CuNN 8.6 were installed to reduce the fitting time. Not every network from previous work could be applied as published; for example, LtCNN is designed for large image patches ($200\times200$) and thus convolutional striding and MaxPooling cannot be applied when patches reach a size of $1\times1$. For LtCNN \citep{lu_hyperspectral_2022}, kernel size and max pooling's strides were reduced as depicted in the Additional data accompanying this work.

\subsection{Classification results}

\begin{figure}[ht]
    \centering
    \includegraphics[width=\linewidth]{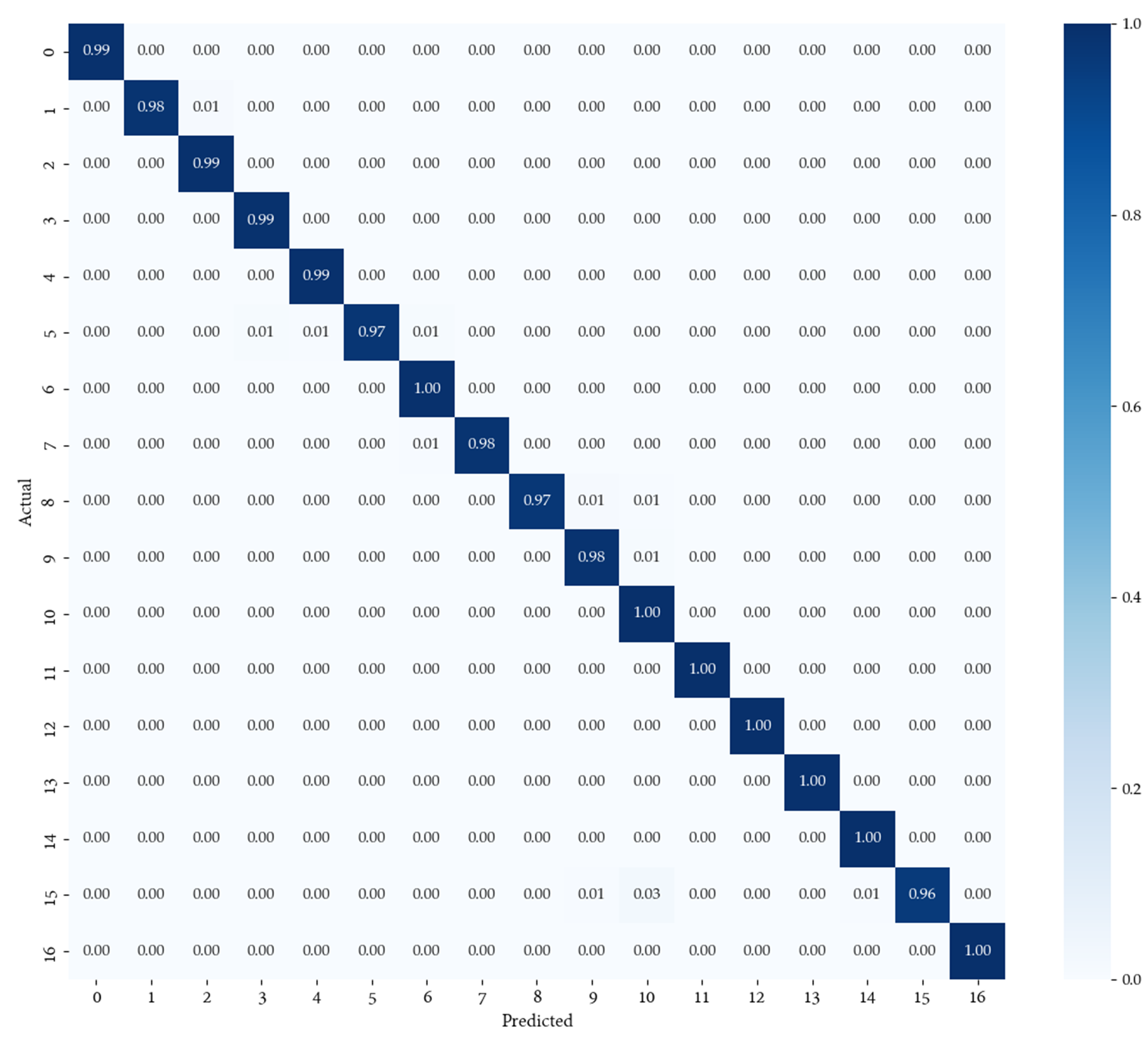}
	\caption{Confusion matrix for classifying red and white varieties altogether. }
	\label{fig:confusion_matrices}
\end{figure}

Table \ref{table:overall_results} shows the overall results of our method in comparison with state-of-the-art networks for classifying HSI datasets. Most of them are considerably unstable due to operating with noisy UAV data, rather than working with satellite imagery. In addition, the second best performing network is \cite{nezami_tree_2020}, which is only the only one checked against UAV datasets for discerning different tree species. Similar to ours, it is also a shallow CNN with only a few layers; however, convolutions are applied in a sequential manner, rather than operating with stacked features extracted from various parallel convolutions. The confusion matrix in \ref{fig:confusion_matrices} shows the OA of the proposed network against any grape variety. Hence, classification over the majority of varieties showed uniform results, with most of them being close to 99\%. Note that these percentages were rounded, and therefore, some of these results are below 99\%, while others are above. When averaged, all these results lead to an OA of $\approx98.8$\%, as shown in Table \ref{table:overall_results}.

\subsection{Training time and network capacity}

\begin{figure}[ht]
    \centering
    \includegraphics[width=\linewidth]{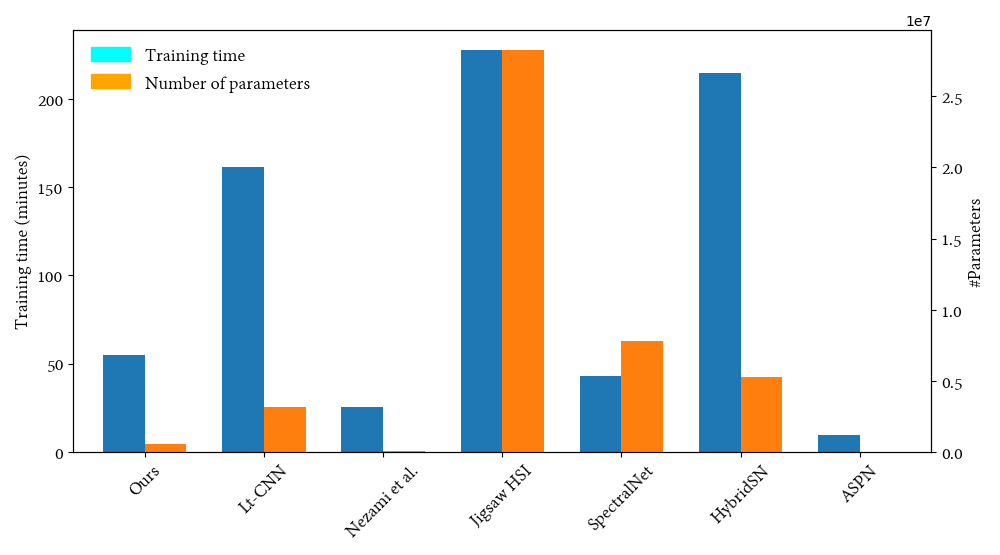}
	\caption{Response time for training the network as well as the number of parameters for every compared network, including ours. }
	\label{fig:time_capacity_networks}
\end{figure}

\begin{figure}[ht]
    \centering
    \includegraphics[width=\linewidth]{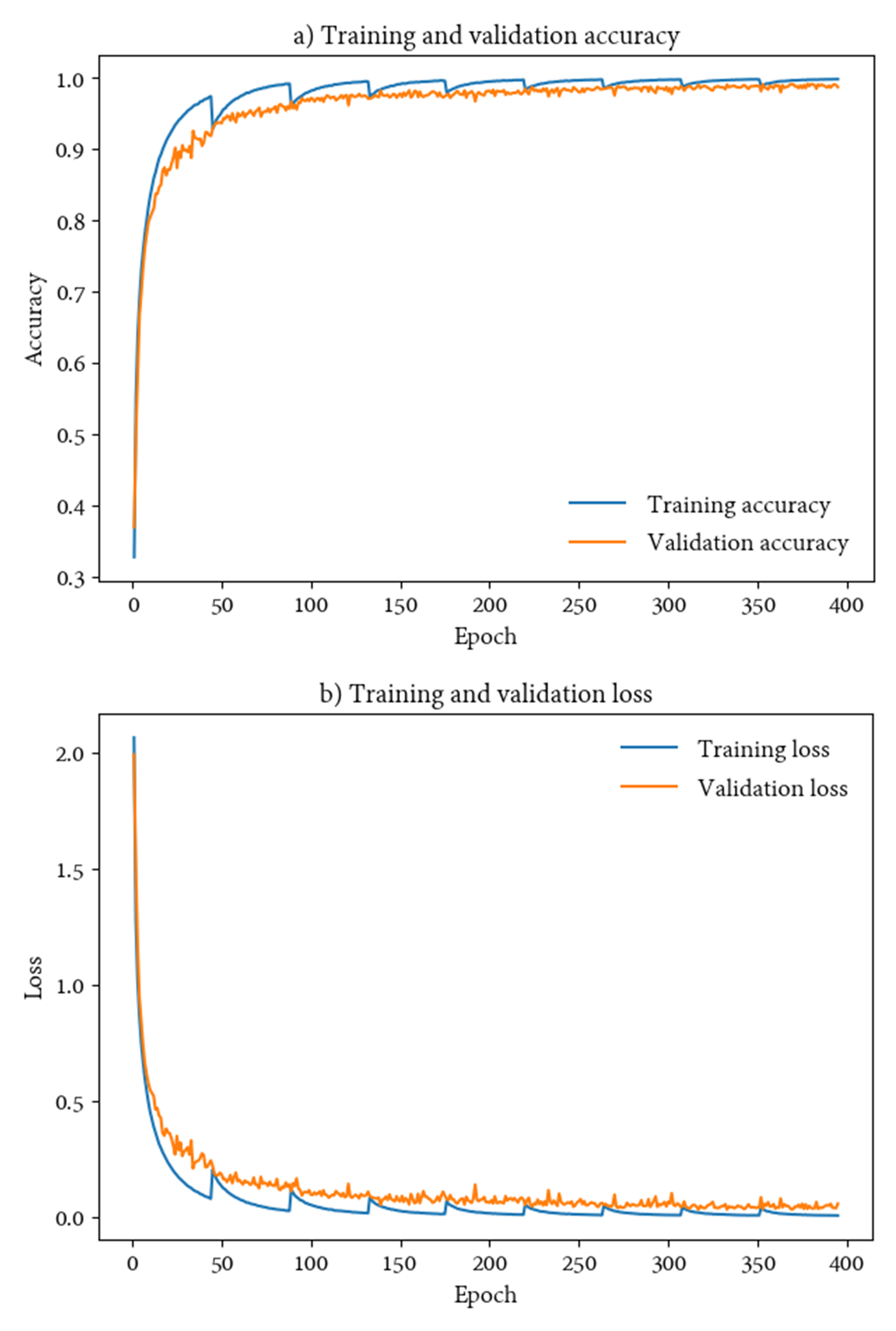}
	\caption{Train and validation accuracy and loss during training. }
	\label{fig:training_history}
\end{figure}

The benchmark on the classification problem is relevant to show that training time is not excessive and that the proposed network does not have much more capacity than the problem warrants. Regarding the capacity, Figure \ref{fig:training_history} shows that training and test signatures are similar, and therefore do not show overfitting nor underfitting behaviours. Furthermore, the training AO and loss worsen as a new dataset is introduced, whereas the validation metrics remain similar. Thus, the network is not memorizing input data. Along with this, the network is only parameterized by nearly 560k parameters, while other state-of-the-art models exceed ten million parameters (see Figure \ref{fig:time_capacity_networks}). The number of parameters is derived from the proposed architectures, using an input of size $23 \times 23 \times 40$. Note that the network of \cite{lu_hyperspectral_2022} was decimated in our experimentation with pooling operations of a lower size than proposed to adapt it to smaller patches. Finally, the response time for training the proposed network is below an hour, whereas others require up to several hours. Note that every available sample was used during training, instead of using strides; otherwise, the training time can be reduced. Further insight into these results is provided by Table \ref{table:overall_results}.

\subsection{Separability}

\begin{figure}[ht]
    \centering
    \includegraphics[width=\linewidth]{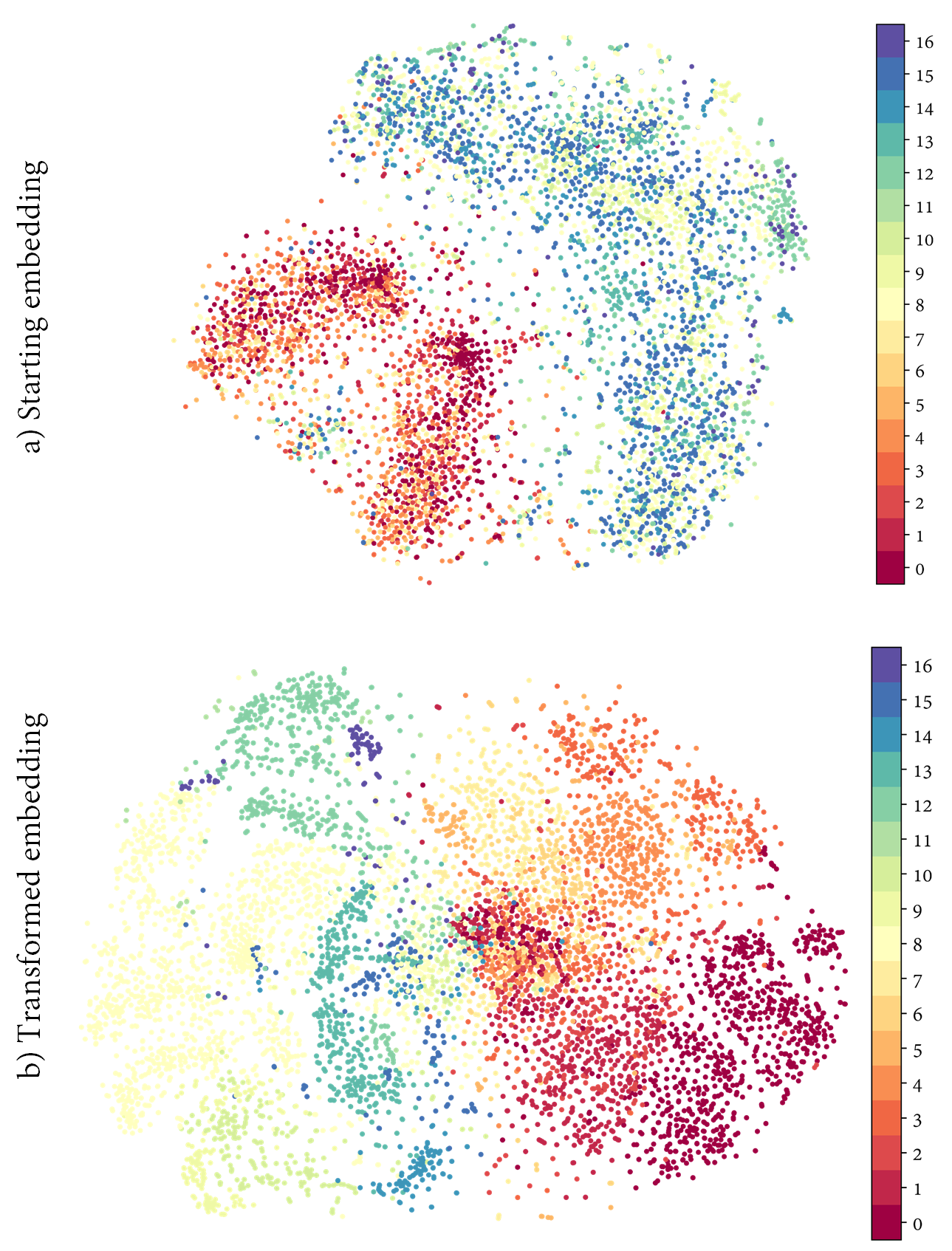}
	\caption{Clustering of samples according to the feature transformation performed by uMAP over 1) the starting hyperspectral features and 2) features extracted by the CNN before transferring it to the final Softmax layer. }
	\label{fig:separability_tsne}
\end{figure}

The output of the proposed network can be assessed in terms of separability by removing the final Dropout and Dense layers. Data is transformed and flattened according to the network's learnt weights. It is subsequently embedded with uMAP \citep{mcinnes_umap_2020} to compress high-dimensionality data into a few features, thus allowing us to visualize the new data representation. The same procedure can be followed over the original data to compare how was the data manifold uncrumpled. As shown in Figure \ref{fig:separability_tsne}, different labels were not perfectly unmixed, although the improvement in comparison to the starting representation is notable. To provide this result, the last densely connected layer was connected to uMAP fitting with $n = 2$; hence, 2592 features were narrowed to two features to represent the embedding in a two-dimensional chart. 

\subsection{Impact of window size}

\begin{figure}[ht]
    \centering
    \includegraphics[width=\linewidth]{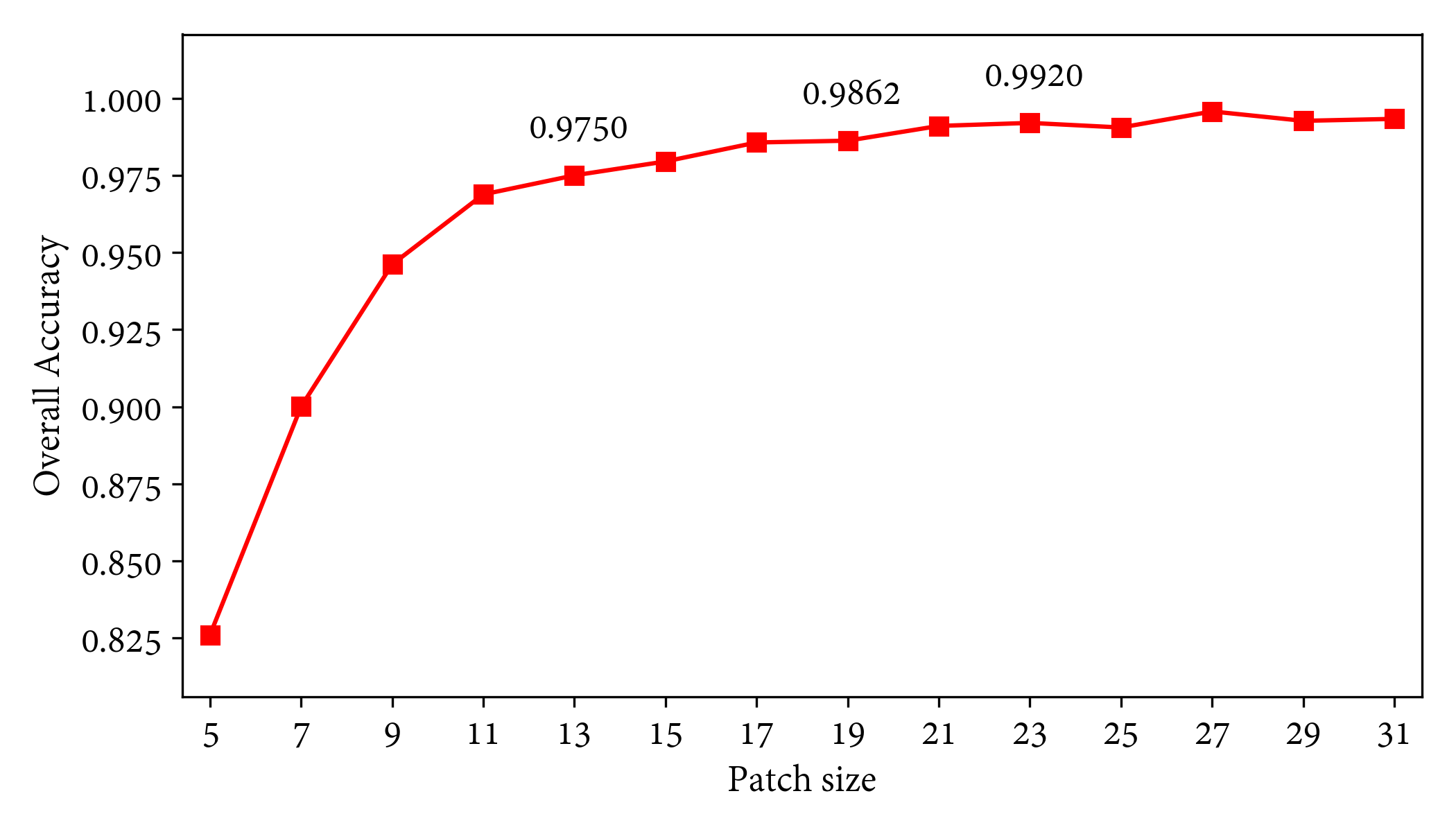}
	\caption{Overall accuracy obtained for patches of different sizes, from 9 to 31. }
	\label{fig:window_size_test}
\end{figure}

\begin{figure}[ht]
    \centering
    \includegraphics[width=\linewidth]{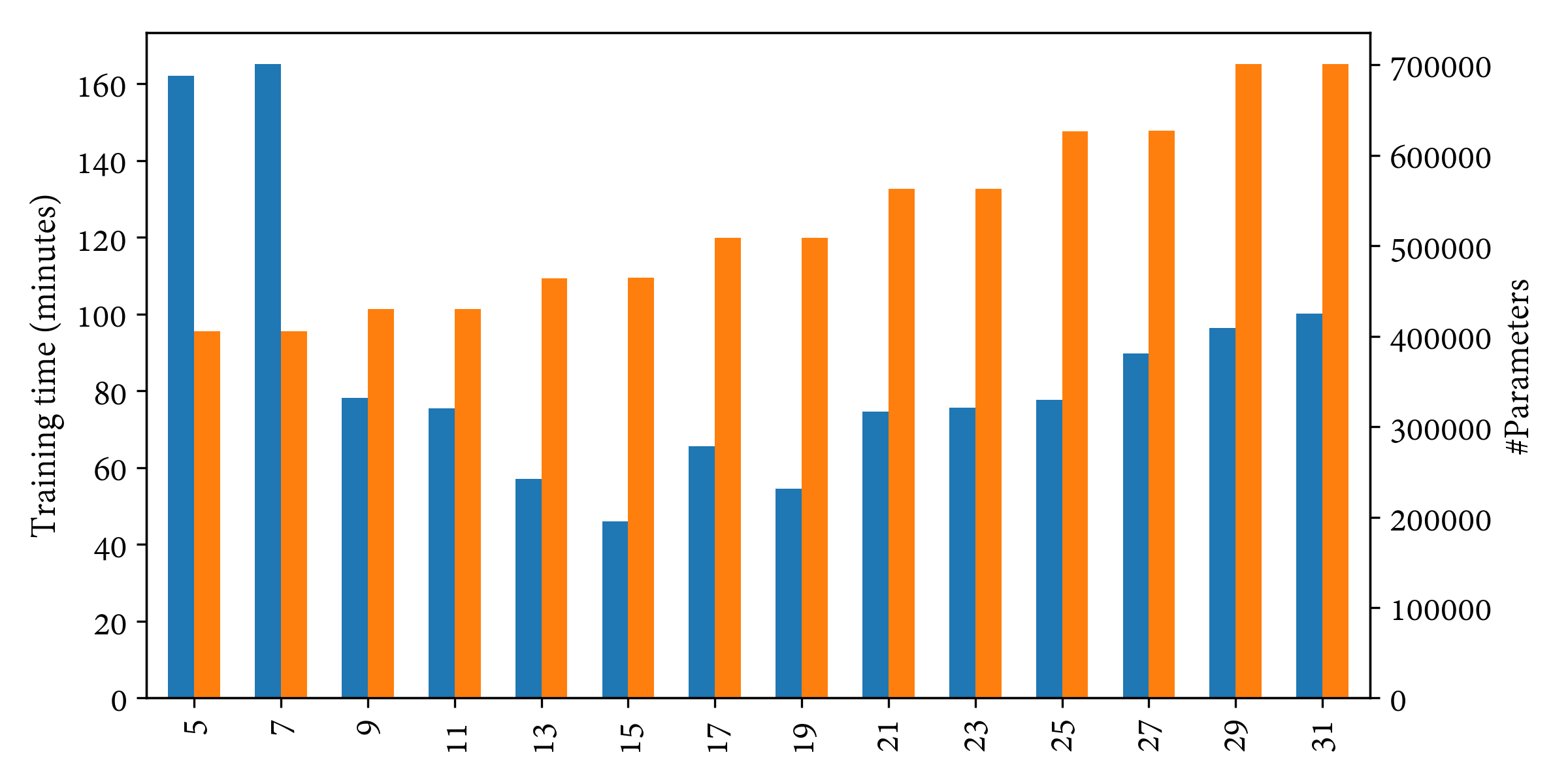}
	\caption{Training time, in minutes, and number of parameters as the window size increases. }
	\label{fig:time_capacity_test}
\end{figure}

The patch size is one, if not the most relevant, parameter concerning the network architecture. Larger patches are assumed to also work as irrelevant spatial features can be zeroed out. However, it also comes at the expense of increasing the training time. On the contrary, lower patches come at the risk of not being sufficient for classifying samples as accurately as done by the proposed network. Figure \ref{fig:window_size_test} shows the whole battery of metrics obtained with patch sizes ranging from 9 to 31. According to the obtained results, patches have been split with dimensionality 23 to balance network capacity and accuracy, despite higher patch size achieving slightly better results. Accordingly, the highest patch size reached an OA of 99.57\%, whereas the lowest reached 82.5\% (size of 9). On the other hand, the selected dimensionality achieves an OA of 99.20\%, thus leveraging network size and capacity. On the other hand, Figure \ref{fig:time_capacity_test} depicts the training time and network size as the patch dimensions increase. The number of training splits was calculated according to the patch size, and therefore, the lowest size had also a lower number of subdivisions. This led to a considerable time bottleneck in patch-wise transformations since they are performed in the Central Processing Unit (CPU). Therefore, it can be observed that the selected size is also intended to leverage training time with capacity and accuracy.

\subsection{Ablation study}

\begin{figure}[ht]
    \centering
    \includegraphics[width=\linewidth]{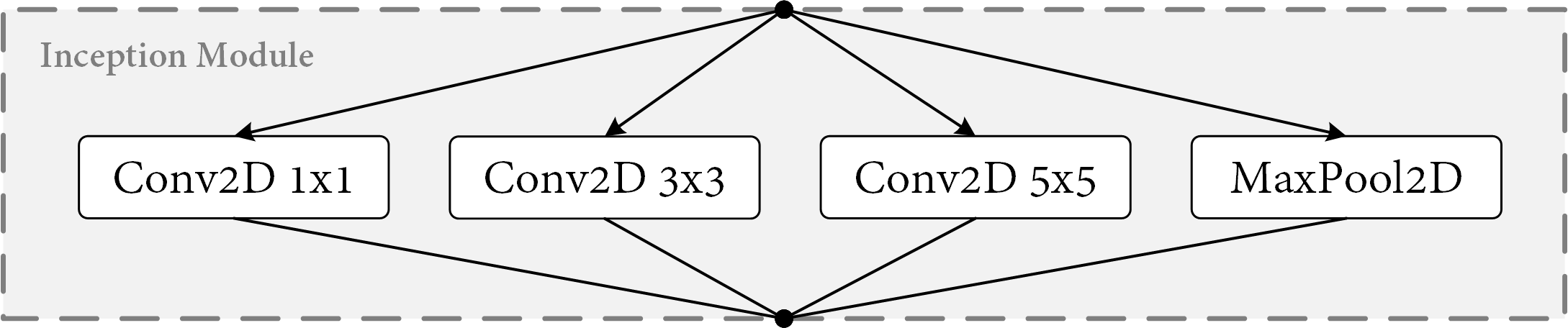}
	\caption{First proposal of Inception block \citep{szegedy_going_2014}.}
	\label{fig:naive_inception}
\end{figure}

The proposed network is intended to be validated in this section by removing and transforming some of the network features, while the rest remain unchanged. The proposed changes are the following:
\begin{enumerate}
    \item Two convolutional layers were included before Part II to extract spectral and spatial features.
    \item Both Inception blocks were modelled using the naïve version \citep{szegedy_going_2014} (see Figure \ref{fig:naive_inception}). The main difference between this architecture and ours is that the former stacks feature maps extracted with different neighbourhood sizes, without downsampling data with $1x1$ convolutions. Therefore, it increases the network capacity and training time.
    \item Only the first Inception block was exchanged by a naïve version, as the one used in the previous experiment.
    \item The spatial attention layer was removed.
\end{enumerate}

The obtained results are shown in Table \ref{table:ablation_results}. Removing the SA layer led to a slight decrease in performance, similar to exchanging the first Inception block. Unsurprisingly, using the naïve version of the Inception layer twice led to a significant performance decrease for every metric, as it kept transforming the spectral dimensionality in a deep layer. The second Inception version also transforms spectral features but rather provides them as an additional layer (concatenation) that can be weighted according to their contribution to the output. Following this reasoning, swapping the first Inception block with the first version of it did not involve a huge performance drop. Improvements to the proposed architecture over the c) variant were very small and therefore may suggest that using either one of them does not offer great changes in the performance. Similar results to this last setup were achieved by removing the spatial attention layer; it did not lead to a significant performance drop, though better and especially, more stable, results were obtained using it. 

\renewcommand{\arraystretch}{1.1}
\begin{table*}
\centering
\caption{Overall results in terms of OA, AA and Kappa coefficient with different CNN schemes. }
\label{table:ablation_results}
\resizebox{\textwidth}{!}{\begin{tabularx}{\textwidth}{|X@{}|*5{c|}}
\toprule
Metric & Ours & a) With initial conv. & b) Naïve Inception & c) Naïve \& Adv. Inception & d) Without SA\\
\cmidrule{1-6}
OA & \textbf{\numberVariance{98.78}{0.15}} & \numberVariance{98.04}{0.11} & \numberVariance{97.87}{0.29} & \numberVariance{98.51}{0.20} & \numberVariance{98.67}{0.23}\\
AA & \textbf{\numberVariance{98.94}{0.09}} & \numberVariance{98.21}{0.06} & \numberVariance{98.09}{0.25} & \numberVariance{98.90}{0.10} & \numberVariance{98.93}{0.11}\\
Kappa ($\kappa$) & \textbf{\numberVariance{99.67}{0.05}} & \numberVariance{99.43}{0.07} & \numberVariance{99.45}{0.08} & \numberVariance{99.58}{0.12} & \numberVariance{99.59}{0.04}\\
f1 & \textbf{\numberVariance{98.78}{0.15}} & \numberVariance{98.04}{0.11} & \numberVariance{97.89}{0.28} & \numberVariance{98.52}{0.20} & \numberVariance{97.66}{0.22}\\
\bottomrule
\end{tabularx}}
\end{table*}
\renewcommand{\arraystretch}{1}

\subsection{Analysis of errors}

\begin{figure}[ht]
    \centering
    \includegraphics[width=\linewidth]{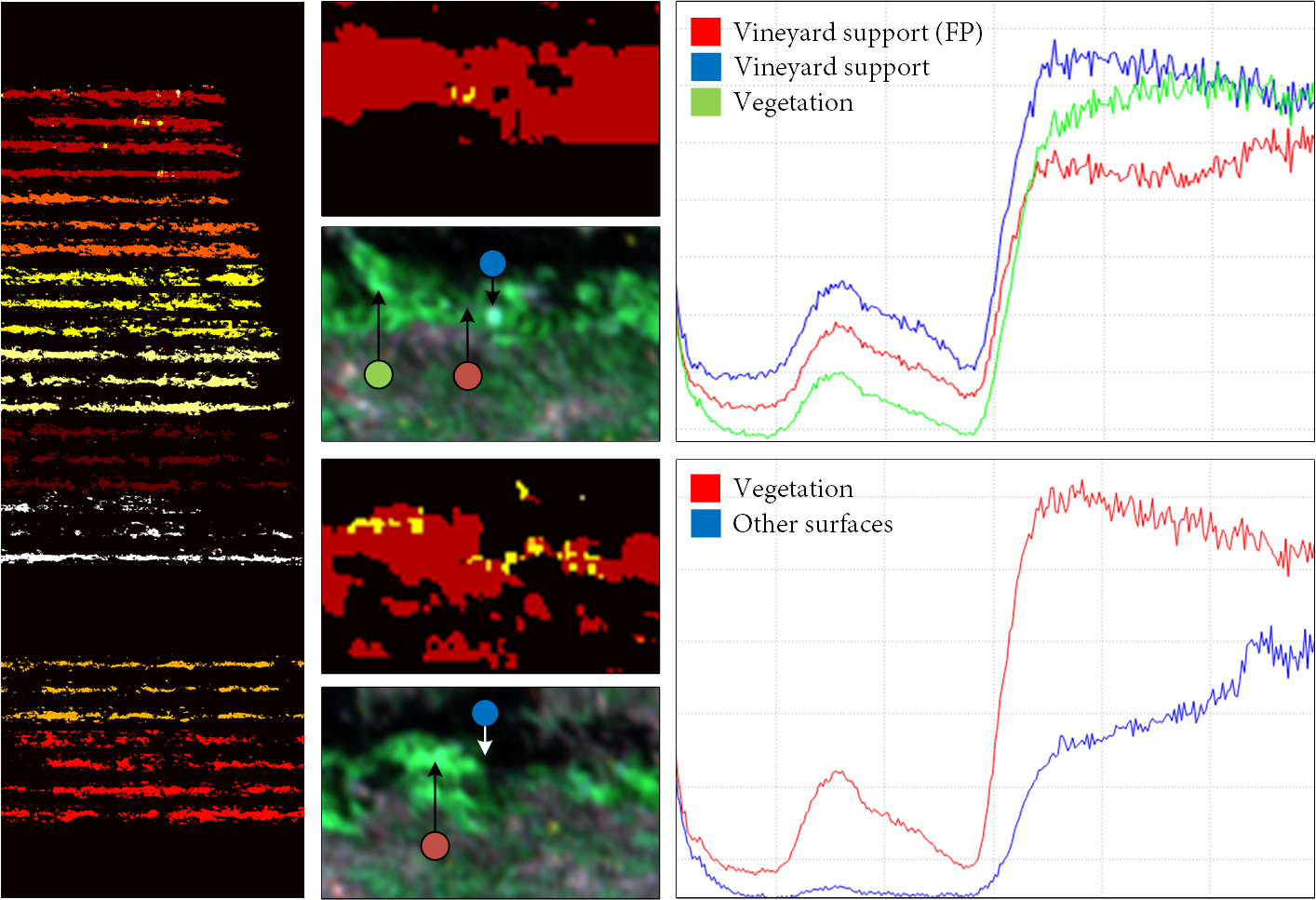}
	\caption{Errors observed in the classification of red varieties, together with the hyperspectral signature of a few samples concerning different surfaces.  }
	\label{fig:spatial_labelling_errors}
\end{figure}

As observed in previous sections, our architecture achieved a high OA and AA. Still, there is a margin for improvement that must be tackled by finding which are the weaknesses in the overall labelling, transformation and classification pipeline. Instead of predicting randomly selected samples, another experiment is to predict every hyperspectral swath sample, thus allowing us to determine where errors take place within the study area. As observed in Figure \ref{fig:spatial_labelling_errors}, these errors are spatially clustered instead of being sparsed over the study area. If these are compared against the RGB mosaic of the hypercubes, errors are observed to belong to 1) small vegetation clusters, mainly proceeding from small vegetation mistakenly labelled as vineyard and 2) samples surrounded by ground or metallic vineyard supports. Note that these are hard to notice during the labelling since they present signatures similar to the target leaves and they are surrounded by vegetation, thus hardening the definition of a geometrical shape for rapidly tagging which is relevant or not. Still, some errors are present in grape samples surrounded by ground and other surfaces since these have a notable impact on the sample's neighbourhood, thus distorting the final probability. Note that every boundary sample is surrounded by vegetation, and therefore, it should be expected that their signature is the fusion of the signatures concerning a target variety and ground. However, it may be not relevant enough to tell apart varieties, thereby leading to more errors on samples heavily surrounded by ground.

\subsection{Training over satellite imagery}

The main shortcoming of compared CNNs focused on satellite imagery is that they obtain a poor performance over the proposed UAV datasets and vineyard varieties. Hyperspectral imagery from UAVs is noisier than satellite observations, with the spectral signature of the latter being more smoothed out. Therefore, previous work did not overcome noise in the classification of vineyards and most of them showed poor performance. Only another architecture tested over UAV samples managed to reach an OA near 80\%, whereas others showing worse results proposed huge networks with millions of parameters, thus overkilling the classification with a large training time. 

The aim of this chapter is to remark on how our network behaves over satellite hyperspectral imagery. Although the network was not designed for this purpose, several tests were launched over publicly available datasets frequently used for comparison. The number of labels in these data ranges from nine to sixteen, and the number of spectral bands also differs from our imaging device. However, FA was fitted to obtain only 40 features per pixel, as proposed for UAV imagery, and so does the architecture of the network. According to the number of samples of each dataset, the batch size was adapted as shown in Table \ref{table:satellite_results}. Every dataset had unlabelled samples which were removed from the training and test datasets to establish a fair comparison with previous work. Unlike our UAV datasets, labels in satellite imagery were imbalanced, with some of them having only a few dozen of examples. Hence, balancing was not applied in satellite datasets to avoid levelling the rest of the classes with others that present scarce examples. 

From the results, it can be observed that the proposed network can also be applied to this kind of imagery despite not being its main area of expertise. All of the datasets converged early to classification metrics over 99\%, and despite not providing state-of-the-art results, these are very close to the current networks that show the best performance. As we did not intend to tune the network for satellite imagery, the learning rate remained as before, and the batch size was scaled according to the number of samples.

\renewcommand{\arraystretch}{1.1}
\begin{table*}
    \centering
    \caption{Classification of hyperspectral imagery from satellite platforms in terms of OA and Kappa coefficient ($\kappa$).}
    \label{table:satellite_results}
    \resizebox{\textwidth}{!}{\begin{tabularx}{\textwidth}{|X@{}|*6{l|}}
    \toprule
    & \multicolumn{3}{c|}{Ours} & \multicolumn{3}{c|}{State-of-the-art}\\
    \cmidrule{1-7}
    Dataset & OA & Kappa ($\kappa$) & Batch size & OA & Kappa ($\kappa$) & Reference work \\
    \cmidrule{1-7}
    \textbf{Pavia university} & \numberVariance{99.97}{0.01} & \numberVariance{99.99}{0.00} & 256 & \numberVariance{100}{0.00} & \numberVariance{100}{0.00} & \cite{moraga_jigsawhsi_2022}\\
    \cmidrule{1-7}
    \textbf{Indian pines} & \numberVariance{99.53}{0.13} & \numberVariance{99.49}{0.14} & 64 & \numberVariance{99.93}{0.07} & \numberVariance{99.89}{0.10} & \cite{ravikumar_hyperspectral_2022}\\
    \cmidrule{1-7}
    \textbf{Salinas valley} & \numberVariance{100}{0.00} & \numberVariance{100.0}{0.00} & 256 & \numberVariance{100}{0.00} & \numberVariance{100}{0.00} & \cite{moraga_jigsawhsi_2022}\\
    \bottomrule
    \end{tabularx}}
\end{table*}
\renewcommand{\arraystretch}{1}

\subsection{Training over fewer examples}

\begin{figure}[ht]
    \centering
    \includegraphics[width=\linewidth]{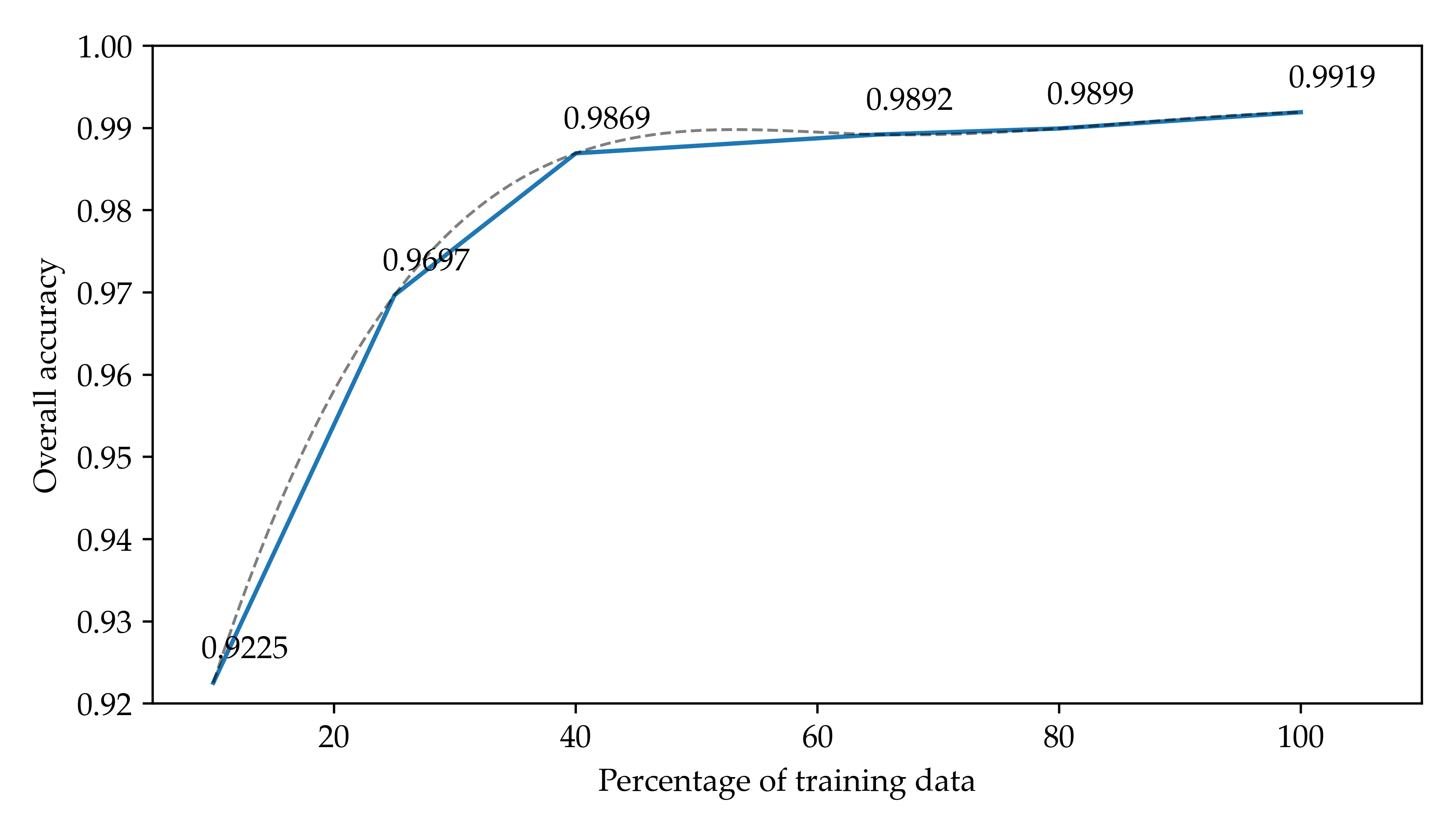}
	\caption{OA observed by training the proposed CNN network with a percentage of the train dataset.  }
	\label{fig:oa_subsampling}
\end{figure}

Another conducted experiment was to train the proposed CNN with a lower amount of information. In this regard, the training was repeated to learn from a percentage of training samples ranging from 10\% to 100\% (of 68\%). In Figure \ref{fig:oa_subsampling}, it can be observed that the OA drastically goes to 92\% with a 10\% of training data, although it is still able to learn relevant features to provide a high OA. It is hypothesized that, as the number of training data increases, the number of learned spatial features is notably higher, whereas lower amounts of information are enough for learning spectral features that enable classifying samples from their neighbourhood. 

\subsection{Transfer learning}

Transfer learning has been widely studied to take advantage of trained networks with a notable capability of separating a variable number of classes. The underlying concept is that a network which has been successfully applied to one case study and has learned relevant features may be applied to another case study with a similar outcome. Nonetheless, it does not necessarily involve training the whole network, which typically has a severe amount of parameters. Instead, some layers that learn more abstract features, presumably the first, are not trained; their weights are preserved and deeper layers are trained to specialize in another application. 

The objective of this section is to carry out several experiments to conclude whether weights learned over the classification of other UAV datasets can be exploited to make the training faster, and even more accurate. This experiment was approached by using the publicly available WHU-Hi HSI datasets for classifying rural materials, including different vegetation crops \citep{zhong_whu-hi_2020}. These datasets are primarily designed for semantic segmentation applications, although the outlined transformation procedure can also be applied to them. It is important to note that these datasets have a lower level of detail (LOD) compared to ours, resulting in ground-truth masks that bear a closer resemblance to satellite imagery. Despite this reduced LOD, they have proven valuable in expediting the training process due to their smaller size. Conversely, the materials depicted in the WHU-Hi datasets exhibit significant dissimilarities from those present in our imagery. The ground-truth masks in these datasets appear smoother due to the lower LOD, leading to a heightened emphasis on spectral features while spatial features contribute less significant information. Consequently, the learned weights from these datasets serve as initial weights, initiating a re-training process focused on acquiring spectral and spatial features from imagery collected with a high LOD.

Table \ref{table:transfer_learning} shows the outcome of this experimentation. Using previously trained weights considerably contributed to improving the metric results, in comparison with the default weight initialization. By default, weights in Keras are initialized so that the variance is guaranteed to be similar across the network layers (Xavier initialization). Note that, not every dataset evenly contributed to improving the results; the weights from Han Chuan and Long Kou datasets seem to contribute better to separate hyperspectral samples. 

\renewcommand{\arraystretch}{1.3}
\begin{table*}
    \centering
    \caption{Classification of hyperspectral imagery with weights learnt from WHU-Hi datasets and default Keras weights. }
    \label{table:transfer_learning}
    \resizebox{\textwidth}{!}{\begin{tabularx}{\textwidth}{|X@{}|*3{l|}l|l|l|}
    \toprule
    & \multicolumn{3}{c|}{Previously trained weights} & \multicolumn{3}{c|}{Default weights}\\
    \cmidrule{1-7}
    Dataset & OA & Kappa ($\kappa$) & f1 & OA & Kappa ($\kappa$) & f1\\
    \cmidrule{1-7}
    \textbf{Han Chuan} & \numberVariance{99.10}{0.07} & \numberVariance{99.75}{0.01} & \numberVariance{99.10}{0.07} & 
    \multirow{3}{*}{\numberVariance{98.78}{0.15}} & 
    \multirow{3}{*}{\numberVariance{99.67}{0.05}} & 
    \multirow{3}{*}{\numberVariance{98.78}{0.15}}\\
    \textbf{Hong Hu} & \numberVariance{98.79}{0.13} & \numberVariance{99.68}{0.03} & \numberVariance{98.79}{0.12} & & &\\
    \textbf{Long Kou} & \numberVariance{99.09}{0.13} & \numberVariance{99.73}{0.07} & \numberVariance{99.09}{0.13} & & &\\
    \bottomrule
    \end{tabularx}}
\end{table*}
\renewcommand{\arraystretch}{1}

\section{Discussion and conclusions}
\label{sec:discussion}

The proposed network presents a CNN model that uses current state-of-the-art procedures applied to classify grapevine varieties. Spatial-attention layers were proved to enhance the results, whereas Inception blocks were checked to discuss which one could provide better results over hyperspectral imagery. Note that the majority of CNN works in the literature are oriented towards RGB imagery, and therefore, their transition to hypercubes is not as trivial. Also, only a few works have addressed the classification of UAV-based HSI, since it is considerably noisier. Despite this network being proposed for phenotyping applied to grapevines, it was also tested over standard satellite HSI datasets which are frequently used to establish fair comparisons. The results showed that samples from satellite imagery were also classified with notable accuracy (over 99.7\%). In contrast, most of the compared models achieved poor results over UAV-based imagery ($\sim81$\% at most). Our training time and the number of parameters were also lower than most of the compared classification models; some of them are composed of millions of parameters, whereas ours solely consists of $\sim560K$ parameters. 

Another explored step, which may be the most relevant, is the preprocessing of reflectance. First, hypercubes are composed of a huge number of bands that cannot be processed by CNNs in a reasonable response time. Furthermore, it hardens the fitting stage and most of these bands may be either redundant or irrelevant to the classification. A significant number of feature reduction and transformation methods were checked, which led to Factor Analysis as the one providing better results in an automatic pipeline. The latter concept is relevant since other methods may be capable of providing better clusters at the expense of demanding the number of different materials collected in hyperspectral imagery. Using FA, hypercubes with 270 bands were narrowed to only 40 features, thus lowering the number of trained parameters and response time. 

Regardless of the high OA and AA, the analysis of errors also depicted some clustered samples that were mislabelled. If checked against false-colour RGB imagery from hypercubes, most of them were low-vegetation labelled as grape varieties or samples surrounded by other surfaces which hardened the classification. 

As a future work, we would like to further extend this network to integrate as many more varieties as possible. It also ought to be explored whether a different growth stage has some effects on the classification. Hence, proving this would enable this work to be used over any study area at any stage of the year as long as it comprises some of the varieties over which the model has been trained. Also, a considerable effort is being made to collect information about crops receiving grants from national and European funds. For instance, it should be inferred whether a crop is abandoned or not to make a decision on whether these funds are granted. Similarly to this, estimating the area of vineyard rows, even the harvested varieties, could also help in this and other decision-making processes. Finally, the labelling may greatly benefit from using the Digital Elevation Model as collected by the surveying UAV, in order to avoid some labelling errors and speed up this manual task. 

\section{Acknowledgments}
\label{sec:acknowledgments}

This result has been partially supported by the Spanish Ministry of Science, Innovation and Universities via a doctoral grant to the first author (FPU19/00100), as well as a grant for researching at the University of Trás-os Montes e Alto Douro (EST22/00350). It has also been partially supported by the project “DATI—Digital Agriculture Technologies for Irrigation Efficiency”, PRIMA—Partnership for Research and Innovation in the Mediterranean Area (Research and Innovation activities), financed by the states participating in the PRIMA partnership and by the European Union through Horizon 2020.

\printcredits

\bibliographystyle{cas-model2-names}

\bibliography{references}

\end{document}